\documentclass[lettersize,journal]{IEEEtran}
\usepackage{amsmath,amsfonts}
\usepackage{algorithmicx}
\usepackage{algcompatible}
\usepackage{algorithm}
\usepackage{array}
\usepackage[caption=false,font=normalsize,labelfont=sf,textfont=sf]{subfig}
\usepackage{textcomp}
\usepackage{stfloats}
\usepackage{url}
\usepackage{verbatim}
\usepackage{graphicx}
\usepackage{cite}
\usepackage{booktabs}
\usepackage{multirow}
\usepackage{bbding}
\usepackage{hwemoji}
\usepackage{tcolorbox}
\usepackage{xcolor}
\usepackage{mdframed}
\usepackage{wasysym}    
\usepackage{pifont}
\usepackage{algpseudocode} 
\usepackage{algorithm}

\definecolor{refblue}{RGB}{31, 119, 180}
\definecolor{refred}{RGB}{214, 39, 40}
\definecolor{refgreen}{RGB}{44, 160, 44}

\newcommand{\cmark}{\ding{51}} 

\newcommand{\spl}[1]{{\color{blue} #1}}

\begin{document}

\title{Memory-Augmented Multimodal Large Language Models for Small Object Understanding in Streaming Aerial Videos}

\author{Penglei Sun$^{1}$, 
Yehua Huang$^{1}$,
Zhuoli Tao$^{2}$,
Xiang Li$^{1}$,
Runwei Guan$^{1}$,
Yaoxian Song$^{3}$,
Kaiyong Zhao$^{4}$,\\
Henghui Ding$^{5}$,
Bo Han$^{6}$~\IEEEmembership{Senior Member,~IEEE},
Yang Yang$^{1}$~\IEEEmembership{Fellow,~IEEE},
Xiaowen Chu$^{1}$\textsuperscript{\Envelope}~\IEEEmembership{Fellow,~IEEE}
\thanks{\Envelope~Corresponding author.}
\thanks{$^{1}$The authors are with Information Hub, The Hong Kong University of Science and Technology (Guangzhou), Guangzhou, China, Emails: 
    {\tt\small \{psun012,yhuang704,xli906\}@connect.hkust-gz.edu.\\cn}, 
    {\tt\small \{runwayrwguan,yyiot,xwchu\}@hkust-gz.edu.cn}.
    }
\thanks{$^{2}$The author is with University of Freiburg, Freiburg, German, Emails: 
    {\tt\small zhuoli.tao@students.uni-freiburg.de.
    }
}
\thanks{$^{3}$The author is with Hangzhou City University, Hangzhou, China, Emails: 
    {\tt\small songyaoxian@westlake.edu.cn.
    }
}
\thanks{$^{4}$The author is with XGRIDS, Shenzhen, China, Emails: 
    {\tt\small kyzhao@xgrids.com.
    }
}
\thanks{$^{5}$The author is with Fudan University, Shanghai, China, Emails: 
    {\tt\small hhding@fudan.edu.cn.
    }
}
\thanks{$^{6}$The author is with Department of Computer Science,
Hong Kong Baptist University, Hong Kong, China, Emails: 
    {\tt\small bhanml@comp.hkbu.edu.hk.
    }
}
}

\markboth{Journal of \LaTeX\ Class Files,~Vol.~14, No.~8, August~2021}%
{Shell \MakeLowercase{\textit{et al.}}: A Sample Article Using IEEEtran.cls for IEEE Journals}


    \maketitle

\begin{abstract}

Language-guided aerial perception aims to understand user-specified tiny targets in complex unmanned aerial vehicle (UAV) scenes. 
In real UAV deployment, the UAV must respond while it flies, so such perception runs in an online streaming manner, where frames arrive sequentially and the model responds to each one without access to future frames. 
However, applying current Multimodal Large Language Models (MLLMs) to this setting raises two challenges. 
First, targets viewed from the air are often tiny, yet the visual compression in existing MLLMs treats all regions equally and discards their fine-grained details. 
Second, understanding a continuous stream requires past-frame context, yet retaining the entire history is infeasible on resource-constrained onboard hardware, whereas discarding it causes the target to drift or disappear. 
We address the tiny object and streaming challenges from both data and method perspectives. 
From the data perspective, we present \textbf{DroneEyes}, the \textbf{first} pixel-level and open-vocabulary referring-segmentation dataset for tiny aerial targets, comprising $2,140$ high-definition videos and $176,623$ pairs across Object Description and Referring Expression tasks, with dense per-frame masks. 
From the method perspective, we propose \textbf{SkyAnchor}, an MLLM with two designs to the above challenges: a Semantics-Aware Token Router that preserves small-target under a reduced visual-token budget, and a Hierarchical Memory Bank that keeps the target consistently understood on streams. 
Experiments on DroneEyes show that SkyAnchor outperforms existing MLLMs on both object description (by $2\times$ in GPT score) and referring expression (by $15.3\%$ in region-and-contour accuracy). 
On the unseen SkyFind benchmark, SkyAnchor improves over the previous domain-specific SOTA by $25.9\%$ in IoU@0.5 and $95.5\%$ in IoU@mean on the unseen maritime test split, demonstrating strong cross-domain generalization.
Finally, we deploy SkyAnchor on a real UAV platform, where our optimization pipeline achieves a $3.05\times$ speedup over vanilla PyTorch, demonstrating its practical value for UAV surveillance and operator assistance.
{Our project can be seen in \underline{\textit{\url{https://sites.google.com/view/skyanchor/main-page}}.}}
\end{abstract}


\section{Introduction}
\label{sec:introduction}

\begin{figure}
    \centering
    \includegraphics[width=1.0\linewidth]{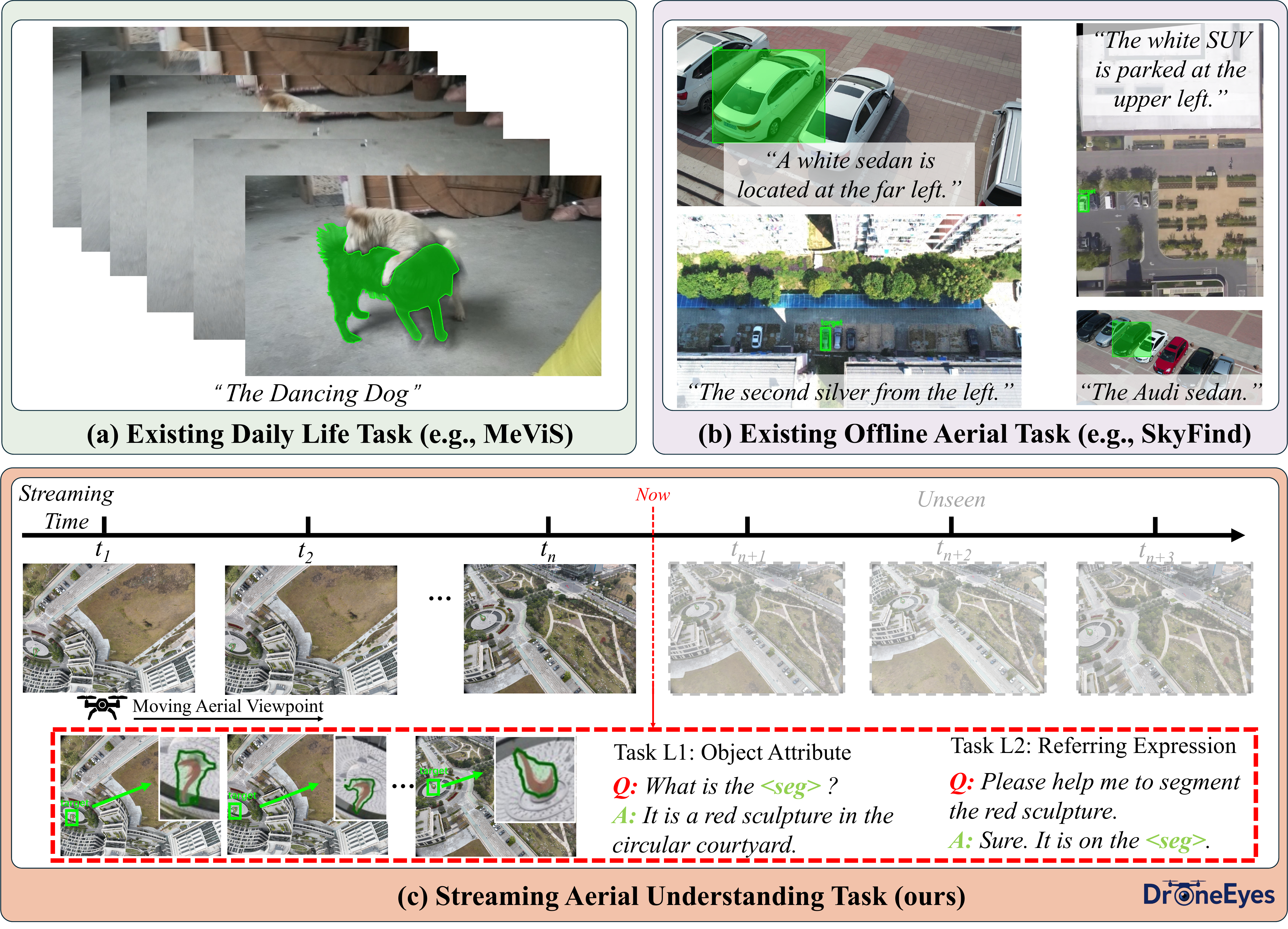}
    \caption{Comparison of DroneEyes with existing daily-life and aerial offline tasks. 
    \spl{\textbf{(a)} Daily-life datasets (e.g., MeViS) feature large objects irrelevant to aerial perception. 
    \textbf{(b)} Existing aerial datasets (e.g., SkyFind) target small objects using bound box assume offline access to complete clips or static images. 
    \textbf{(c)} DroneEyes try to understand small objects using segmentation in streaming aerial video.
    The model must respond at each step without access to future frames, while accurately understanding small objects.
    }}
    \label{fig:intro}
\end{figure}

Unmanned aerial vehicles (UAVs) are increasingly deployed in surveillance, search-and-rescue, and infrastructure inspection~\cite{huang2025small,kaufmann2023champion,maninis2018video,sun20243d}. 
To assist operators in these applications, UAVs are expected to follow natural-language instructions and ground the referred targets in streaming aerial video, e.g., \textit{``Find the white truck.''}~\cite{weyler2024phenobench}. 
Since the UAV must respond while it flies, this perception cannot be performed offline over a complete clip, but must run in an online streaming manner. 
This setting poses two intrinsic challenges. 
First, targets viewed from the air are often tiny, occupying only a small portion of high-resolution frames~\cite{rozantsev2016detecting}, which demands pixel-accurate perception to separate them from cluttered background. 
Second, frames are captured and consumed sequentially, so the model must respond to each frame the moment it arrives without accessing future frames~\cite{plummer2020revisiting}, while the resource-constrained onboard device cannot cache the entire visual history of an ever-growing stream.

Existing datasets do not provide the annotations required for this task, as illustrated in Fig.~\ref{fig:intro}. 
First, daily-life datasets such as MeViS and LaMOT~\cite{ding2023mevis,li2025lamot} feature large objects that are far from the tiny targets seen from aerial viewpoints, and they draw their referring expressions from a closed set of common categories. 
Second, existing aerial datasets such as SkyFind and AerialMind~\cite{wang2026skyfind,chen2026aerialmind} capture the bird's-eye view but offer coarse supervision, typically bounding boxes or frame-level captions, without pixel-accurate masks that delineate small targets from cluttered background. 

To fill this gap, we construct DroneEyes, the \textbf{first} pixel-level and open-vocabulary referring-segmentation dataset for tiny aerial targets. 
It comprises $2,140$ high-definition drone videos and $176,623$ pairs across Object Description and Referring Expression tasks. 
Unlike existing aerial datasets, DroneEyes preserves all videos at their native high resolution and annotates dense per-frame segmentation masks over an open vocabulary along continuous flight videos, so that the fine-grained shape of each tiny target is delineated from the background throughout the stream. 
The dataset is organized into two task levels from perception to end-to-end grounding. 
Level 1 (L1) provides a target region and requires the model to describe what is in that region, fostering fine-grained perception of small objects. 
Level 2 (L2) provides a natural-language referring expression and requires the model to segment the specified target, probing end-to-end referring segmentation.

Beyond data, although recent multimodal large language models (MLLMs) have shown strong video understanding capabilities, the two challenges above also carry over to the model side when they are applied to streaming aerial video~\cite{liu2026event}. 
First, some targets occupy only small portions of high-resolution frames~\cite{rozantsev2016detecting}, leaving most visual tokens on background regions~\cite{sun2025city}. 
Full-token processing is accurate but costly on edge devices, while existing compression in general MLLMs is efficient but degrades tiny-object information.
For instance, Qwen series~\cite{qwen2025qwen25technicalreport} applies a fixed-ratio token merging that reduces all spatial regions equally.
It will treat a small vehicle the same as a large building, so that small targets are merged into surrounding background tokens.
Second, understanding a continuous video stream requires past-frame context, yet retaining the entire history is infeasible for streams, whereas discarding it causes the target to drift or disappear.
For instance, SAM 3~\cite{carion2025sam} accumulates per-frame memory features whose storage scales linearly with sequence length.
This strategy \spl{cannot} sustain long-term identity.

To address the above challenges of MLLMs, we propose SkyAnchor, an MLLM designed for small-object understanding in streaming aerial video. 
For the first challenge, we introduce a Semantics-Aware Token Router that preserves target-relevant information through semantics-aware weighted aggregation under a fixed token budget.
For the second challenge, we design a Hierarchical Memory Bank that decouples temporal information into two complementary layers. 
The Semantic Memory layer stores object-level appearance features to preserve target identity over the streaming term, while the Tracking Memory layer caches recent spatial states to support short-term mask propagation. 
The two layers work jointly to keep a stable understanding of the target on streams.

Experiments on DroneEyes validate the effectiveness of SkyAnchor against both general-purpose MLLMs and specialist segmentation models. 
On the L1 object description task, SkyAnchor achieves $2\times$ performance over the strongest general-purpose baselines across metrics and demonstrating clear advantages in small-object understanding. 
On the L2 referring segmentation task, SkyAnchor-3B reaches \spl{best region-and-contour accuracy ($\mathcal{J}\&\mathcal{F}$)}, surpassing the existing models by $15.3 \%$ despite using far fewer parameters. 
To assess cross-domain generalization, we further evaluate SkyAnchor on the SkyFind benchmark without any fine-tuning.
On the unseen maritime test split, SkyAnchor surpasses the previous domain-specific SOTA by $25.9\%$ in IoU@0.5 and $95.5\%$ in IoU@mean, demonstrating strong robustness under aerial domain shifts.
Finally, we deploy SkyAnchor on a real UAV platform with NVIDIA Jetson AGX Orin 64GB. 
With our optimization pipeline, the system achieves a $3.05\times$ speedup over vanilla PyTorch.
It demonstrates cloud-free online monitoring on the target platform~\cite{ghosh2023react,yang2019multiple}.

In conclusion, this paper makes the following contributions:

\begin{itemize}
    \item From the data perspective, we construct \textbf{DroneEyes}, a large-scale aerial streaming video dataset comprising $2,140$ high-definition drone videos and $176,623$ pairs with dense per-frame segmentation masks. 
    To our knowledge, DroneEyes is the \textbf{first} aerial streaming understanding dataset with the pixel-level annotation.

    \item From the method perspective, we propose \textbf{SkyAnchor}, an MLLM that introduces a Semantics-Aware Token Router for preserving fine-grained details of tiny targets under the same budget and a Hierarchical Memory Bank for consistent understanding on streams, enabling efficient small-object understanding in streaming aerial video.

    \item On DroneEyes, SkyAnchor with only 3B parameters achieves state-of-the-art results on both L1 object description and L2 referring segmentation, outperforming all MLLMs. 
    On the SkyFind benchmark without any fine-tuning, SkyAnchor achieves  improvements over the previous domain-specific SOTA on the unseen maritime test split, with relative gains of $25.9\%$ in IoU@0.5 and $95.5\%$ in IoU@mean, demonstrating strong cross-domain generalization.
    
    \item From the deployment perspective, we develop a optimization pipeline and validate SkyAnchor on real-world surveillance videos captured by the UAV. 
    Our optimization pipeline achieves $3.05\times$ speedup over vanilla PyTorch, confirming practical applicability for onboard inference without cloud offloading.

\end{itemize}

The rest of this paper is structured as follows. 
We begin with a review of related work in Section~\ref{sec:related_work}. 
Section~\ref{sec:dataset} introduces the DroneEyes dataset with the construction engine. 
Section~\ref{sec:method} formulates the problem and introduces the proposed framework SkyAnchor. 
Section~\ref{sec:experiment} presents the experiments and analyses. 
Finally, Section~\ref{sec:conclusion} concludes the paper and outlines future directions.



\begin{table*}[htbp]
  \centering
  \small
  
  \caption{Comparative Dataset for Video Segmentation. We compare the existing video
  segmentation dataset from venues.
  Seg. means segmentation task.
  Granularity indicates the spatial precision of the annotations.
  Pixel-level denotes dense per-frame segmentation masks and Bbox-level denotes bounding-box annotations.
  Expr. denotes instance level referring expressions for visual grounding. Bboxes are detection annotations without language. QA pairs serve video question answering.
  Avg O/I Ratio is the arithmetic mean of $(area_{object} / area_{image})  \times 100\%$ across all annotations in the dataset.
  }
  \label{tab:data_comparsion}
  \resizebox{1.0\textwidth}{!}{%
  
  \begin{tabular}{@{}c|c|c|c|c|c|c|c|cc|c|c|c@{}}
\toprule
\multirow{2}{*}{\textbf{Dataset}}                                  & \multirow{2}{*}{\textbf{Venue}} & \multirow{2}{*}{\textbf{\begin{tabular}[c]{@{}c@{}}Aerial\\ Perspective\end{tabular}}} & \multirow{2}{*}{\textbf{Granularity}} & \multirow{2}{*}{\textbf{Modality}} & \multirow{2}{*}{\textbf{\begin{tabular}[c]{@{}c@{}}Open\\ Vocab.\end{tabular}}} & \multirow{2}{*}{\textbf{\begin{tabular}[c]{@{}c@{}}Tiny \\ Target\end{tabular}}} & \multirow{2}{*}{\textbf{\begin{tabular}[c]{@{}c@{}}Avg\\ O/I Ratio\end{tabular}}} & \multicolumn{2}{c|}{\textbf{Tasks}}                             & \multirow{2}{*}{\textbf{Videos}} & \multirow{2}{*}{\textbf{Expression}} & \multirow{2}{*}{\textbf{Resolution}} \\ \cmidrule(lr){9-10}
                                                                   &                                 &                                                                                        &                                       &                                    &                                                                                 &                                                                                  &                                                                                   & \textbf{Seg.}                  & \textbf{QA}                    &                                  &                                      &                                      \\ \midrule
Ref-Youtube-VOS~\cite{seo2020urvos}          & ECCV 2020                       &                                                                                        & Pixel-Level                           & Video                              & \cmark                                                           &                                                                                  &     18.31 \%                                                                              & \cmark          &                                & 3,978                            & 15K Expr.                            & $\le$ 2.7K                           \\
ShareGPT4Video~\cite{chen2024sharegpt4video} & NIPS 2024                       &                                                                                        & Frame-Level                                     & Video                              & \cmark                                                           &                                                                                  & -                                                                                 &                                & \cmark          & 40K                              & 40K Captions                         & $\le$ 720p                           \\
MeViS~\cite{ding2025mevis}                   & TPAMI 2025                      &                                                                                        & Pixel-Level                           & Video                              & \cmark                                                           &                                                                                  & 1.4\%                                                                             & \cmark          &                                & 2,006                            & 33K Expr.                            & $\le$ 720p                           \\
JTD-UAV~\cite{wang2025jtd}                   & CVPR 2025                       & \cmark                                                                  & BBox-Level                            & Video                              &                                                                                 & \cmark                                                            & 0.21\%                                                                            &                                & \cmark          & 1,328                            & 3K QA pairs                          & 1080p                                \\
LaMOT~\cite{li2025lamot}                     & ICRA 2025                       &                                                                                        & BBox-Level                            & Video                              & \cmark                                                           &      \cmark                                                                            & 0.63\%                                                                            &                                & \cmark          & 62                               & 145 Expr.                            & $\le$ 720p                           \\
AerialMind~\cite{chen2026aerialmind}         & AAAI 2026                       & \cmark                                                                  & BBox-Level                            & Video                              & \cmark                                                           & \cmark                                                            & 0.25\%                                                                            &                                &                                & 93                               & 24.6K Expr.                          & $\le$ 720p                           \\
SkyFind~\cite{wang2026skyfind}               & TPAMI 2026                      & \cmark                                                                  & BBox-Level                            & Image                              & \cmark                                                           & \cmark                                                            & 0.76\%                                                                            &                                &                                & 35K Images                       & 1M Expr.                             & $\le$ 4K                             \\ \midrule
\textbf{DroneEyes (ours)}                                          & \textbf{-}                      & \textbf{\cmark}                                                         & \textbf{Pixel-Level}                  & \textbf{Video}                     & \textbf{\cmark}                                                  & \textbf{\cmark}                                                   & 0.24\%                                                                            & \textbf{\cmark} & \textbf{\cmark} & 2,140                            & 176K pairs                           & \textbf{$\le$ 4K}                    \\ \bottomrule
\end{tabular}
}
\end{table*}

\section{Related Work}
\label{sec:related_work}

\subsection{Referring Video Object Segmentation}

Referring video object segmentation (RVOS) segments target objects in videos given natural language expressions. 
Existing methods follow two main lines. 
The first builds upon foundation segmentation models. 
VISA~\cite{yan2024visa}, Sa2VA~\cite{yuan2025sa2va} and UniPixel~\cite{liu2026unipixel}  uses a language model to reason on a single frame and hands the mask to SAM 2~\cite{ravi2024sam} for propagation.
The second pursues end-to-end architectures that jointly handle language grounding and mask prediction.
VideoGLaMM~\cite{munasinghe2025videoglamm} produces spatially grounded masks alongside language outputs.
GLUS~\cite{lin2025glus} and VRS-HQ~\cite{gong2025devil} integrate reasoning within a single large language model. InstructSeg~\cite{wei2025instructseg} and further unify diverse referring and segmentation tasks at the pixel level. 
However, these methods are designed for daily life level videos with relatively large objects and do not address the challenges specific to aerial scenarios, including small targets, dense distributions, large viewpoint changes, and the requirement for continuous streaming processing.

\subsection{MLLMs for Aerial Understanding}

MLLMs have recently been adapted to remote sensing and aerial domains. 
GeoChat~\cite{kuckreja2024geochat}  SkySenseGPT~\cite{luo2024skysensegpt} and EarthGPT~\cite{zhang2024earthgpt}  improve perception by enriching training data or broadening input modalities to SAR and infrared. 
SkyFind~\cite{zhang2024earthgpt} introduces a million-scale UAV referring expression benchmark and reveals that current MLLMs struggle with small-target localization under heavy background clutter, while JTD-UAV\cite{wang2025jtd} combines MLLMs with tracking for joint localization and intent description in thermal sequences. 
Both remain limited to box-level output or static grounding without pixel-level video segmentation. 
Our work addresses this gap by performing referring segmentation on aerial video streams with hierarchical memory and token compression tailored for small objects at high resolution.

\section{DroneEyes Dataset}
\label{sec:dataset}

\begin{figure}
    \centering
    \includegraphics[width=1.0\linewidth]{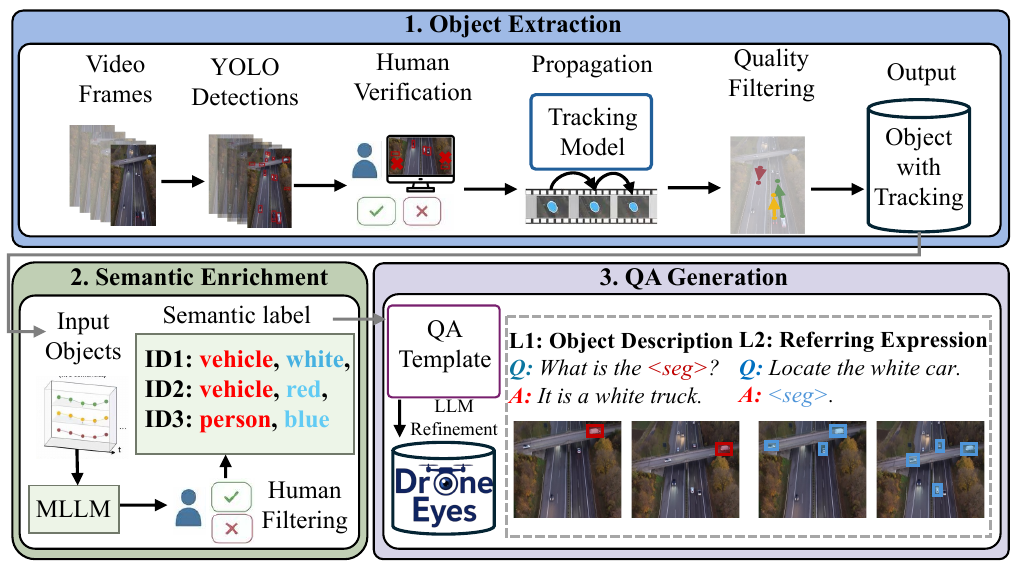}
    \caption{The data label pipeline for DroneEyes.
    \textbf{(1) Object Extraction}. The candidate objects are detected with human verification, followed by mask propagation.
    \textbf{(2) Semantic Enrichment}. We use VLM to enrich attribute descriptions via context-aware visual prompting.
    \textbf{(3) QA Generation}. Two-level QA pairs are generated with LLM-based linguistic refinement.}
    \label{fig:data_label}
\end{figure}


\subsection{Data Collection and Diversity}
We curate a dataset of $2,140$ high-definition drone videos sourced from the open-access platform Pexels\footnote{\url{https://www.pexels.com/}}. Unlike ground-level datasets, aerial visual understanding depends on fine-grained spatial details. Therefore, we preserve all videos at their native resolutions, typically 1080p or 4K, to retain the visual cues of small objects.

To ensure data diversity, the collected videos cover various real-world topologies including urban cityscapes, highway traffic, agricultural fields, and natural terrain under different illumination conditions, altitudes, and camera trajectories such as top-down, oblique, and orbital. This collection provides a diverse distribution of common aerial targets, including vehicles, pedestrians, and infrastructure, supporting the training of grounding-capable vision-language models.

\subsection{Data Engine}
Annotating dense spatial grounding and complex understanding pairs for drone videos is labor-intensive due to the massive number of small targets. 
To scale up dataset construction, we develop an automated data engine that couples foundation segmentation models with large language models. 
The pipeline consists of three main stages: extracting consistent spatial-temporal object trajectories, and enriching these bounding boxes with semantic descriptions to construct difficulty-progressive tasks, as shown in Fig~\ref{fig:data_label}.

\subsubsection{Object Extraction}
To establish consistent object trajectories, we implement a hybrid pipeline combining automated detection and human verification. We first utilize a YOLO-based detector to generate initial bounding box proposals on video keyframes. Human annotators then verify and filter these proposals to ensure high-precision initialization. Using these verified boxes as spatial prompts, we employ Segment Anything Model 3 as our foundation tracker. The tracker propagates segmentation masks bidirectionally through the video to maintain consistent object identities. To construct the final ground truth, we enforce a strict quality control mechanism by retaining only the tracking outputs with high confidence scores. A final manual sampling inspection guarantees the overall accuracy and reliability of the generated annotations.

\subsubsection{Semantic Enrichment}
Raw spatial trajectories lack semantic information. We leverage a Vision-Language Model to generate descriptive attributes for each tracked object. To overcome the recognition challenge of small aerial targets, we apply context-aware visual prompting. Specifically, we extract object regions with expanded contextual boundaries and apply explicit visual markers. This spatial guidance allows the MLLM to accurately produce category labels and fine-grained visual details. The resulting data bridges the gap between geometric tracking and semantic understanding.

\subsubsection{Data Generation}
Aerial visual reasoning requires understanding objects at multiple scales, from identifying the attributes of a single tiny vehicle to inferring the functional layout of an entire intersection. To systematically evaluate and train these capabilities, we build upon the semantically enriched annotations to design a two level generation strategy. 
This strategy requires complex visual understanding:

\begin{itemize}
    \item \textbf{Level 1 (L1): Object Description.} The model identifies an object specified by a bounding box region and describes its visual attributes.
    \item \textbf{Level 2 (L2): Referring Expression.} The model locates a target object based on a natural language description and provides its exact spatial coordinates.
\end{itemize}

To further improve linguistic diversity, we apply an LLM refinement stage. 
A primary challenge during this text polishing process is preventing the LLM from hallucinating or modifying the original spatial coordinates. 
We address this issue by decoupling the coordinates from the text using an abstract placeholder mechanism during the LLM prompt stage. 
The exact bounding box values are restored only in the final post-processing step.

\subsection{Data Statistics}

\begin{figure*}
    \centering
    \includegraphics[width=1.0\linewidth]{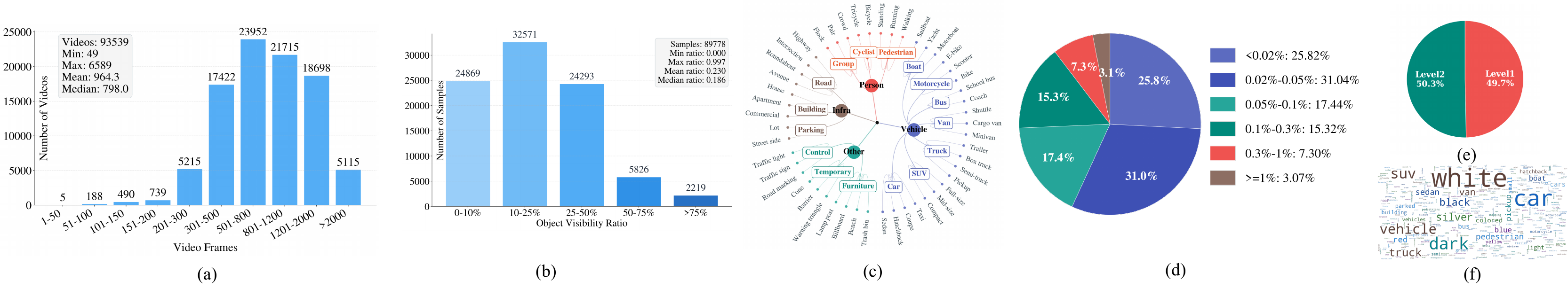}
    \caption{The statistic of DroneEyes.
    (a) Distribution of video frame counts across the dataset.
    (b) Distribution of object visibility ratios across annotation samples.
    (c) Hierarchical taxonomy of object categories in the dataset.
    (d) Distribution of object area ratios relative to the image area.
    (e) Split between Level 1 and Level 2 annotations in the dataset.
    (f) Word cloud of query text, with word size reflecting term frequency.
    }
    \label{fig:data_stats}
\end{figure*}

As summarized in Table~\ref{tab:data_comparsion}, existing video benchmarks each cover only part of what real aerial perception requires, whether limited to a single task or relying on heavily compressed footage. 
Prior aerial video datasets in particular target geometric perception such as box-level detection, where annotations are closed-set category labels with bounding boxes. 
DroneEyes instead provides open-vocabulary, language-driven supervision.
Rather than coarse bounding boxes, it offers pixel-level segmentation masks, and it is the first to pair referring segmentation with question answering on the same continuous aerial stream, with masks and answers aligned to individual frames. 
In Table~\ref{tab:data_comparsion}, Our DroneEyes dataset has an average object-to-image ratio of only $0.24 \%$, lower than LaMOT ($0.63 \%$) and SkyFind ($0.76 \%$), making it one of the more demanding benchmarks for small object drone visual understanding.
Such pixel-level annotation is far more informative than bounding boxes for tiny targets, and together with native uncompressed streams ranging from 1080p to 4K, it is what preserves their structural details and enables accurate pixel-level mask generation in complex aerial sensing environments.
DroneEyes dataset comprises \textbf{176,623} pairs generated from \textbf{2,140} valid drone videos (videos containing no detectable objects were excluded). Fig.~\ref{fig:data_stats} summarizes the more data statistics for DroneEyes.


\section{SkyAnchor Method}
\label{sec:method}

SkyAnchor answers the language query and segments the target across a drone video stream with a vision-language backbone and a mask decoder (Sec.~\ref{sec:overview}). 
Building on this, we design two components: a hierarchical memory bank (Sec.~\ref{sec:memory}) and a semantics-aware token router (Sec.~\ref{sec:token_router}). 
We further deploy it with an edge inference schedule (Sec.~\ref{sec:edge_deployment}).
We summarize the overall procedure in Alg.~\ref{algor:skyanchor}.

\begin{figure}
    \centering
    \includegraphics[width=1.0\linewidth]{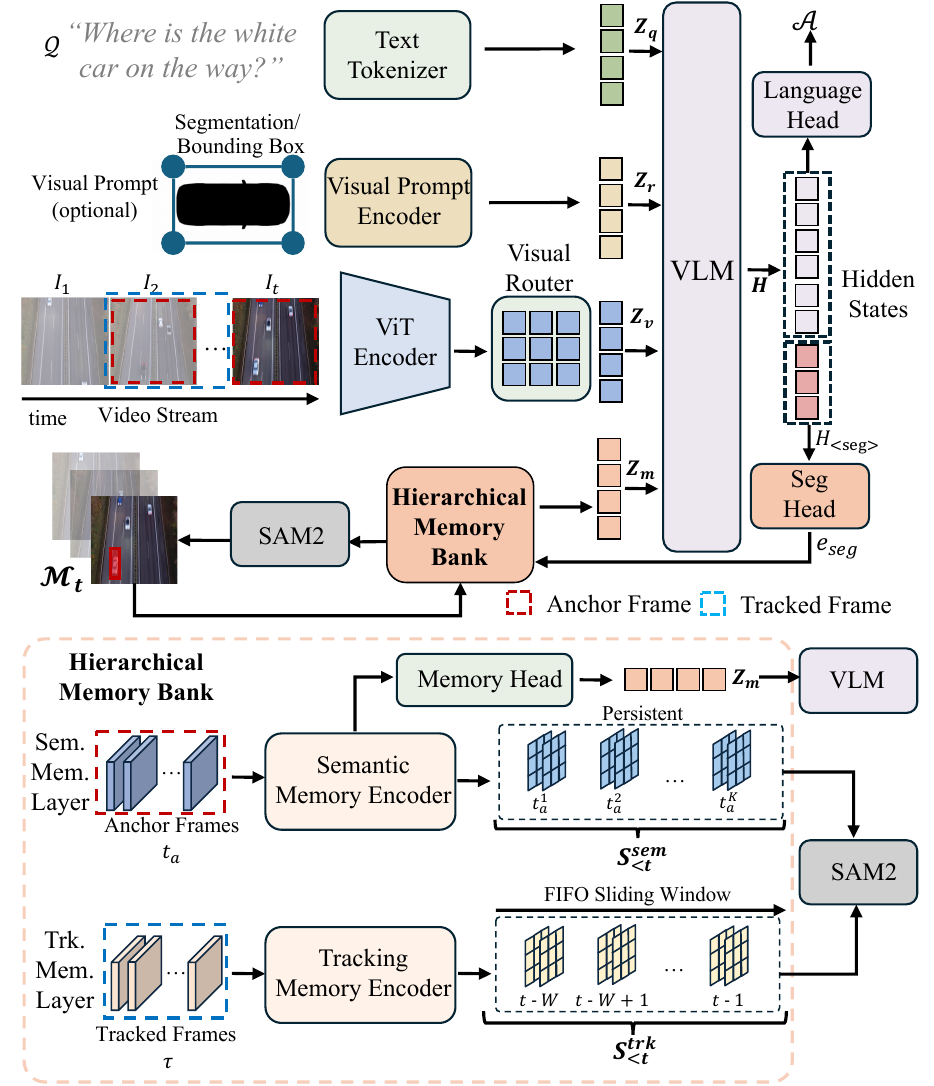}
    \caption{
Overview of SkyAnchor. 
\textbf{(A) Model Overview}: $\mathcal{Q}$, optional visual prompt, and $\{I_1,\dots,I_t\}$ are encoded into $\mathbf{Z}_q$, $\mathbf{Z}_r$, $\mathbf{Z}_v$; together with $\mathbf{Z}_m$, the VLM produces $\mathbf{H}$, from which the Language Head outputs $\mathcal{A}$ and the Seg Head outputs $\mathbf{e}_{\text{seg}}$ to drive SAM2 for $\mathbf{M}_t$.
\textbf{(B) Hierarchical Memory Bank}: the Semantic Memory Encoder registers persistent anchors $\mathcal{S}^{\text{sem}}_{<t}$ from anchor frames (\textcolor{red}{red dashed}) and exports $\mathbf{Z}_m$ to the VLM via the Memory Head. 
The Tracking Memory Encoder caches recent frames (\textcolor{blue}{blue dashed}) in a FIFO window $\mathcal{S}^{\text{trk}}_{<t}$. 
\textbf{(C) Semantics-Aware Token Router}: the Visual Router compresses dense ViT tokens into compact $\mathbf{Z}_v$ by merging background tokens while retaining small-target details.
    }
    \label{fig:method}
\end{figure}





\subsection{Model Overview}
\label{sec:overview}

\subsubsection{Task Statement}
We address the task of \textbf{online referring video object segmentation} in drone-captured aerial footage. Given a continuous video stream $\{I_t\}_{t=1}^{\infty}$ and a natural language query $\mathcal{Q}$, the system must produce a segmentation mask $\mathbf{M}_t$ and the answer $\mathcal{A}$ for the query for each frame upon arrival:
\begin{equation}
    \mathbf{M}_t, \mathcal{A} = \mathcal{F}(I_t, \mathcal{Q}, \mathcal{S}_{<t}),
    \label{eq:online_formulation}
\end{equation}
where $\mathcal{S}_{<t}$ denotes the accumulated memory state. This online, causal formulation is essential for real-world drone applications requiring low-latency streaming results.

Our model realizes $\mathcal{F}$ through a unified architecture comprising two principal components: (i)~a vision-language model (VLM) backbone $\mathcal{F}_{\text{vlm}}$ that reasons over visual and textual inputs, and (ii)~a mask prediction decoder $\mathcal{F}_{\text{dec}}$ that produces per-frame segmentation masks. To handle arbitrarily videos, the memory is explicitly decoupled into permanent semantic anchors $\mathcal{S}^{\text{sem}}$ and sliding tracking states $\mathcal{S}^{\text{trk}}$ (detailed in Sec.~\ref{sec:memory}).

\subsubsection{Vision-Language Backbone}
We build upon Qwen2.5-VL. At each anchor frame $t_a$ (triggered every $N$ frames), the visual encoder extracts dense spatial features from the image. To prevent these from overwhelming the LLM's attention computation, our Token Router (Sec.~\ref{sec:token_router}) dynamically compresses them into a compact visual token sequence $\mathbf{Z}_v$. The query $\mathcal{Q}$ is tokenized into text embeddings $\mathbf{Z}_q$. Additionally, a reference encoder provides spatial prompt embeddings $\mathbf{Z}_r$, and the semantic memory provides appearance tokens $\mathbf{Z}_m$ from past anchors. 
The backbone jointly processes these inputs:
\begin{equation}
    \mathbf{H} = \mathcal{F}_{\text{vlm}}\!\big([\mathbf{Z}_v;\, \mathbf{Z}_q;\, \mathbf{Z}_r;\, \mathbf{Z}_m]\big).
    \label{eq:vlm_forward}
\end{equation}
Note that $\mathcal{F}_{\text{vlm}}$ runs only at anchor frames to update the semantic understanding; between anchors, mask prediction relies efficiently on the decoder's tracking memory.

\begin{figure}
    \centering
    \includegraphics[width=1.0\linewidth]{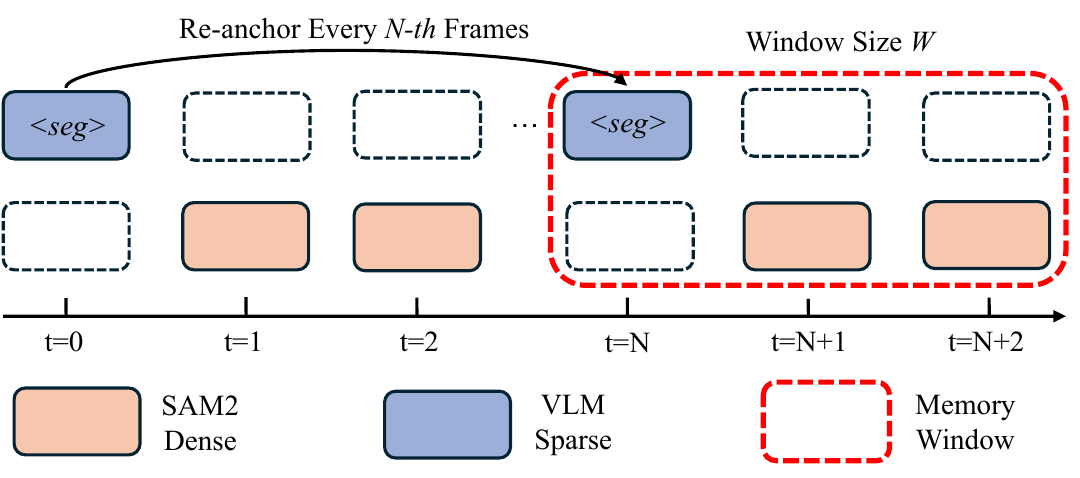}
    \caption{
    \spl{Edge inference schedule of SkyAnchor. 
    The top row is the backbone, which runs only on anchor frames, re-anchoring every $N-th$ frame.
    Dashed boxes mark frames where it is skipped and the cached token is reused. 
    The bottom row is the SAM 2 decoder (orange), which runs on every frame and propagates from the memory bank between anchors. 
    The red dashed box is the tracking memory, a fixed-size FIFO window of size W that slides forward as each new frame enters and the oldest leaves. }
    }
    \label{fig:device}
\end{figure}

\subsubsection{Mask Prediction Decoder}
We adopt SAM2 as the decoder $\mathcal{F}_{\text{dec}}$. Internally, a segmentation head transforms the backbone's hidden states $\mathbf{H}$ at \texttt{<seg>} positions into dense prompt embeddings $\mathbf{e}_{\text{seg}}$. 
The decoder then consumes the memory from Hierarchical Memory Bank based on the $\mathbf{e}_{\text{seg}}$.
The Hierarchical Memory Bank produces combined spatial memory and the SAM2 decoder predict the binary mask:
\begin{equation}
    \mathbf{M}_t = \mathcal{F}_{\text{dec}}( \mathcal{S}^{\text{sem}}_{<t} \cup \mathcal{S}^{\text{trk}}_{<t}).
    \label{eq:dec_forward}
\end{equation}
By querying both the permanent semantic anchors ($\mathcal{S}^{\text{sem}}$) and the recent tracking states ($\mathcal{S}^{\text{trk}}$), this formulation seamlessly unifies long-term identity preservation and short-term multi-frame propagation.

\subsection{Hierarchical Memory Bank}
\label{sec:memory}

A fundamental challenge in online video object segmentation is maintaining object identity across temporal spans. We observe that \textit{``what an object is"} changes slowly, while \textit{``where it is"
} changes rapidly. 
This motivates a two-level hierarchical memory bank equipped with a Semantic Memory Encoder and a Tracking Memory Encoder.

\subsubsection{Semantic Memory Layer}
\label{sec:semantic_memory}
This layer provides persistent identity information to both the VLM branch and the decoder, preventing long-term drift. 
Let $\{t_a^{(i)}\}_{i=1}^{K_t}$ denote all previous anchor frames before the current frame $t$, where $t_a^{(i)} < t$. 
For each anchor frame $t_a^{(i)}$, the Semantic Memory Encoder registers a semantic anchor and materialized differently in the two branches:
\begin{itemize}
    \item \textbf{VLM-side:} When reference masks are available, object-specific appearance features are pooled and projected into a token $\mathbf{z}^{\text{sem}}_{t_a}$. These tokens form the memory embeddings $\mathbf{Z}_m = \{ \mathbf{z}^{\text{sem}}_{t_a} \mid t_a < t \}$ injected into Eq.~\eqref{eq:vlm_forward}.
    \item \textbf{Decoder-side:} The hidden state at $t_a$ acts as a sparse conditioning token. The resulting anchor prediction is encoded into SAM2's native memory format as a 2D spatial feature $\mathbf{m}^{\text{sem}}_{t_a}$. The spatial semantic memory is defined as:
\end{itemize}
\begin{equation}
    \mathcal{S}^{\text{sem}}_{<t}
=
\left\{
\mathbf{m}^{\text{sem}}_{t_a^{(i)}}
\mid
t_a^{(i)} < t
\right\}_{i=1}^{K}.
    \label{eq:sem_memory}
\end{equation}
These continuous representations are stored persistently as conditioning memories. Unlike discretized binary masks, they preserve rich categorical details (e.g., \textit{``silver sedan''}), which is critical for resolving multi-object ambiguities during decoder-side memory attention.

\subsubsection{Tracking Memory Layer}
\label{sec:tracking_memory}
This layer captures fine-grained spatial continuity. 
Between consecutive anchor frames, SAM2 propagates masks, and the Tracking Memory Encoder writes each tracked frame $\tau$ into a 2D tracking state $\mathbf{m}^{\text{trk}}_\tau$.
The tracking memory is defined as:
\begin{equation}
    \mathcal{S}^{\text{trk}}_{<t} = \big\{ \mathbf{m}^{\text{trk}}_\tau \;\big|\; \tau \in [t-W,\, t) \big\}.
    \label{eq:trk_memory}
\end{equation}
Unlike persistent semantic anchors, these are transient, non-conditioning memories consumed by SAM2's memory attention before mask decoding. To bound memory consumption, they are maintained in a FIFO sliding window of size $W$.

\subsection{Semantics-Aware Token Router}
\label{sec:token_router}

Processing high-resolution drone imagery requires a balance between visual detail and computational cost.
We resolve this with a semantics-aware Token Router that dynamically compresses the dense visual tokens into the compact $\mathbf{Z}_v$ consumed by $\mathcal{F}_{\text{vlm}}$.

\subsubsection{Token Merging}
The Token Router partitions the dense spatial token grid into non-overlapping $k \times k$ blocks. 
Within each block, a lightweight scoring network evaluates the semantic salience of each raw token $\mathbf{v}_i$ to produce an importance score $s_i$. The block is then merged into a single output token via softmax-weighted aggregation:
\begin{equation}
    \hat{\mathbf{z}} = \text{Proj}\!\Big(\sum_{i \in \mathcal{B}} \text{softmax}(s_i) \cdot \mathbf{v}_i\Big),
    \label{eq:token_merge}
\end{equation}
where $\text{Proj}(\cdot)$ is a learnable projection. 
The collection of all $\hat{\mathbf{z}}$ across blocks constitutes $\mathbf{Z}_v$. 
The Router uses the same compression ratio for all spatial blocks,
but adjusts the aggregation weights according to their visual
content.

\subsubsection{Routing Loss}
To guide the importance scores toward target-related regions, we guide it with an auxiliary loss supervised by the ground-truth segmentation mask:
\begin{equation}
    \mathcal{L}_{\text{total}} = \mathcal{L}_{\text{vlm}} + \lambda_1 \cdot \mathcal{L}_{\text{seg}} + \lambda_2 \cdot \mathcal{L}_{\text{route}},
    \label{eq:total_loss}
\end{equation}
where $\mathcal{L}_{\text{route}} = \text{BCE}(\mathbf{s},\, \bar{\mathbf{M}}_{\text{gt}})$, with $\mathbf{s}$ being the predicted importance scores and $\bar{\mathbf{M}}_{\text{gt}}$ the ground-truth mask downsampled to the token grid resolution.

\subsection{Edge Deployment}
\label{sec:edge_deployment}

\spl{
We run SkyAnchor on an edge platform as shown in Fig~\ref{fig:device}. 
The multimodal inference runs sparsely in time, while the tracking that emits a mask on every frame stays dense and cheap, so most frames skip the heavy backbone entirely. The backbone $\mathcal{F}_{\text{vlm}}$ is invoked only on anchor frames, spaced $N$ frames apart, where it produces the conditioning token $\mathbf{e}_{\text{seg}}$ that drives the decoder. 
This call is asynchronous, so the per-frame path keeps running on the cached token until the next one is ready and the backbone never stalls the stream. 
Every frame then passes through the SAM~2 path and memory attention propagates the target over the tracking memory $\mathcal{S}^{\text{trk}}_{<t}$, and the decoder $\mathcal{F}_{\text{dec}}$ combines the hierarchical memory with the cached token to emit the mask $\mathbf{M}_t$, so a coherent mask is available even while the backbone is silent.
The tracking memory is a fixed-size FIFO window of capacity $W$, where each frame appends a new state and evicts the oldest once full. On top of this, the backbone, the SAM~2 image encoder and the mask decoder are compiled into static-shape BF16 TensorRT engines, while the per-frame memory attention is captured into a CUDA Graph that replays its many small kernels as one operation.
}

\begin{algorithm}[t]
\caption{SkyAnchor Online Inference}
\label{algor:skyanchor}
\begin{algorithmic}[1]
\Require Video stream $\{I_t\}$, query $\mathcal{Q}$, anchor interval $N$,
         FIFO size $W$, block size $k$
\Ensure  Mask stream $\{\mathbf{M}_t\}$ and answer $\mathcal{A}$
\State \textbf{Init:} $\mathcal{S}^{\text{sem}}\!\leftarrow\!\emptyset$,\;
       $\mathcal{S}^{\text{trk}}\!\leftarrow\!\emptyset$ (capacity $W$),\;
       $\mathbf{e}_{\text{seg}}\!\leftarrow\!\emptyset$
\For{each frame $I_t$}
    \If{$t \bmod N == 1$}
        \State $\hat{\mathbf{z}} \leftarrow \text{Proj}\!\big(\textstyle\sum_{i\in\mathcal{B}}\text{softmax}(s_i)\,\mathbf{v}_i\big)$
        \State $\mathbf{Z}_v \leftarrow \{\hat{\mathbf{z}}\}$
        \State $\mathbf{H} \leftarrow \mathcal{F}_{\text{vlm}}\!\big([\mathbf{Z}_v;\mathbf{Z}_q;\mathbf{Z}_r;\mathbf{Z}_m]\big)$
        \State $\mathbf{e}_{\text{seg}} \leftarrow \text{SegHead}(\mathbf{H})$
        \State $\mathcal{S}^{\text{sem}}.\text{append}(\mathbf{m}^{\text{sem}}_{t})$
    \EndIf
    \State $\mathbf{M}_t \leftarrow \mathcal{F}_{\text{dec}}\!\big(\mathcal{S}^{\text{sem}}_{<t} \cup \mathcal{S}^{\text{trk}}_{<t},\ \mathbf{e}_{\text{seg}}\big)$
    \State $\mathcal{S}^{\text{trk}}.\text{append}(\mathbf{m}^{\text{trk}}_t)$
    \If{$|\mathcal{S}^{\text{trk}}| > W$}
        \State pop oldest from $\mathcal{S}^{\text{trk}}$
    \EndIf
    \State \textbf{yield} $\mathbf{M}_t,\ \mathcal{A}$
\EndFor
\end{algorithmic}
\end{algorithm}


\section{Experiment}
\label{sec:experiment}

\begin{table*}
\centering
\caption{Quantitative comparison of object description and referring small object segmentation performance on the DroneEyes.
Existing MLLM baselines are evaluated using their native offline inference mode, whereas SkyAnchor performs online streaming inference.
Seg. LLM denotes the MLLM with segmentation cabilities.
FT denotes the model finetuned on the DroneEyes train set.
}
\label{tab:l1}
\begin{tabular}{@{}c|ccc|ccc|ccc@{}}
\toprule
\multirow{2}{*}{\textbf{Model}}                                  & \multirow{2}{*}{\textbf{Type}} & \multirow{2}{*}{\textbf{Source}} & \multirow{2}{*}{\textbf{Size}} & \multicolumn{3}{c|}{\textbf{L1 - Object Description}}    & \multicolumn{3}{c}{\textbf{L2 - Referring Expression}}                                \\ \cmidrule(l){5-10} 
                                                                 &                                &                                  &                                & \textbf{ROUGE-L} & \textbf{METEOR} & \textbf{GPT Scores} & \textbf{$\mathcal{J}$} & $\mathbf{\mathcal{F}}$ & $\mathbf{\mathcal{J}\&\mathcal{F}}$ \\ \midrule
Gemini-3.0                                                       & General MLLM                   & Closed                           & -                              & 18.40            & 19.10           & 1.21                & 1.51                   & 4.01                   & 2.76                                \\
GPT-5                                                            & General MLLM                   & Closed                           & -                              & 17.44            & 16.10           & 1.61                & 0.87                   & 2.62                   & 1.75                                \\
GPT-4o                                                           & General MLLM                   & Closed                           & -                              & 18.91            & 21.11           & 1.42                & 0.75                   & 2.04                   & 1.39                                \\
Claude-Sonnet-4.5                                                & General MLLM                   & Closed                           & -                              & 16.36            & 20.50           & 1.47                & 0.54                   & 1.37                   & 0.95                                \\
Kimi-K2.5~\cite{team2026kimi}              & General MLLM                   & Closed                           & -                              & 18.13            & 22.40           & 1.43                & 4.14                   & 7.85                   & 6.00                                \\
Qwen-3.6-A3B~\cite{qwen36plus}             & General MLLM                   & Open                             & 35B                            & 17.61            & 21.80           & 1.46                & 13.12                  & 24.63                  & 18.88                               \\
Gemma-4                                                          & General MLLM                   & Open                             & 31B                            & 19.06            & 18.85           & 1.30                & 16.21                  & 30.10                  & 23.16                               \\ \midrule
SAM3~\cite{carion2025sam}                  & Seg. MLLM                      & Open                             & 1B                             & -                & -               & -                   & 12.67                  & 14.04                  & 13.35                               \\
VISA~\cite{yan2024visa}                    & Seg. MLLM                      & Open                             & 7B                             & -                & -               & -                   & 10.05                  & 10.43                  & 10.24                               \\
VideoGlaMM~\cite{munasinghe2025videoglamm} & Seg. MLLM                      & Open                             & 7B                             & -                & -               & -                   & 11.34                  & 15.54                  & 13.44                               \\
InstructSeg~\cite{wei2025instructseg}      & Seg. MLLM                      & Open                             & 3B                             & -                & -               & -                   & 27.16                  & 36.59                  & 31.87                               \\
GLUS~\cite{lin2025glus}                    & Seg. MLLM                      & Open                             & 7B                             & -                & -               & -                   & 20.50                  & 25.55                  & 23.03                               \\
VRS-HQ~\cite{gong2025devil}                & Seg. MLLM                      & Open                             & 7B                             & -                & -               & -                   & 27.19                  & 36.26                  & 31.73                               \\
Sa2VA~\cite{yuan2025sa2va}                 & Seg. MLLM                      & Open                             & 4B                             & -                & -               & -                   & 21.81                  & 29.10                  & 25.45                               \\
Sa2VA~\cite{yuan2025sa2va}                 & Seg. MLLM                      & Open                             & 8B                             & -                & -               & -                   & 21.36                  & 30.20                  & 25.78                               \\
Sa2VA~\cite{yuan2025sa2va}                 & Seg. MLLM                      & Open                             & 26B                            & -                & -               & -                   & 27.61                  & 38.49                  & 33.05                               \\
UniPixel~\cite{liu2026unipixel}            & Seg. MLLM                      & Open                             & 3B                             & -                & -               & -                   & 23.69                  & 31.83                  & 27.76                               \\
UniPixel~\cite{liu2026unipixel}            & Seg. MLLM                      & Open                             & 7B                             & -                & -               & -                   & 27.96                  & 36.57                  & 32.26                               \\ \midrule
Sa2VA-FT~\cite{yuan2025sa2va}              & Seg. MLLM                      & Open                             & 4B                             & -                & -               & -                   & 24.21                  & 31.41                  & 27.81                               \\
UniPixel-FT~\cite{liu2026unipixel}         & Seg. MLLM                      & Open                             & 3B                             & -                & -               & -                   & 25.16                  & 33.51                  & 29.34                               \\
InstructSeg-FT~\cite{wei2025instructseg}   & Seg. MLLM                      & Open                             & 3B                             & -                & -               & -                   & 28.61                  & 37.76                  & 33.19                               \\ \cmidrule(r){1-9}
SkyAnchor (ours)                                                 & Unified                        & Open                             & 3B                             & \textbf{45.40}   & \textbf{46.31}  & \textbf{3.22}       & \textbf{30.96}         & \textbf{45.23}         & \textbf{38.10}                      \\ \bottomrule
\end{tabular}
\end{table*}

\subsection{Experimental Setup}

\subsubsection{Datasets}
We evaluate our proposed SkyAnchor comprehensively across two distinct domains. 
\begin{itemize}[]

\item For the primary dynamic video scenarios, we utilize our proposed \textbf{DroneEyes} dataset, which provides unconstrained drone trajectories and pixel-level mask annotations. The dataset is split into training, validation, and testing sets with a ratio of 8:1:1.

\item To assess cross-domain generalization on static aerial imagery, we further benchmark our model on the recently released \textbf{SkyFind}~\cite{wang2026skyfind} dataset, following its official validation and test splits. 
Its validation set ($5,000$ samples) shares the same distribution as the training set ($331,364$ samples), whereas its test set ($16,546$ samples) consists of unseen maritime scenarios and thus introduces a clear domain shift. We perform no training on SkyFind and evaluate our model directly on both the validation and test split.
\end{itemize}

\subsubsection{Evaluation Metrics}
For the DroneEyes-L1 subset, we evaluate the generated descriptions using standard natural language generation metrics, including METEOR, ROUGE-L and GPT scores.
For GPT scores, we employ a GPT-based score (0--5 scale) using GPT-5.2 as the judge, which evaluates category accuracy, attribute correctness, spatial reasoning, and descriptive fluency given the original video frames as visual context as the existing setting~\cite{liu2023visual}.
For DroneEyes-L2 subset, we adopt the widely acknowledged standard metrics: the region similarity ($\mathcal{J}$), the contour accuracy ($\mathcal{F}$), and their average ($\mathcal{J}\&\mathcal{F}$). 
For referring image segmentation on SkyFind, we report the standard Intersection over Union at a threshold of 0.5 (IoU@$0.5$) and the mean IoU (IoU@mean), following the SkyFind~\cite{wang2026skyfind} setting.
To better characterize performance under different aggregation protocols, we report both macro-average and micro-average results.
The macro-average treats the validation and test splits equally, while the micro-average weights each sample according to the split size.

\subsubsection{Implementation Details}
\label{sec:implementation}
Our architecture is built upon the pre-trained Qwen2.5-VL (3B parameters) as the vision-language backbone and SAM~2 as the mask prediction decoder. 
The model is optimized using AdamW with an initial learning rate of $2 \times 10^{-5}$ and a cosine annealing scheduler. The LLM backbone is fine-tuned via LoRA ($r{=}128$, $\alpha{=}256$). 
The training is conducted on 8 H800 GPUs with per-device batch size 4. 
For the Token Router, the merge size is set to 2, the maximum patch size to 960, and the auxiliary routing loss weight $\lambda_1 = \lambda_2 = 1$. The router is trained with a separate learning rate of $10^{-4}$, while the SAM~2 decoder uses $5 \times 10^{-6}$.
In the benchmark experiment, the models are inferenced on the H800 GPU.

\subsubsection{Comparative Methods}

To  evaluate our models, we benchmark it against diverse state-of-the-art methods aligned with different tasks. 
\begin{itemize}[]
    \item For the Object Description task (DroneEyes Level 1 Task), we compare SkyAnchor against general MLLMs.
    For all baseline models, we sample frames from the input video and provide the target bounding box as normalized text coordinates $[x_{\min}, y_{\min}, x_{\max}, y_{\max}]$ in the prompt, asking the model to describe the specified object. 
    \item For the Referring Expression task (DroneEyes Level 2 Task), we evaluate spatial localization acuity against existing general and segmentation MLLMs. 
    For general MLLMs, we provide the query and frames and ask them to output the target bounding box as normalized text coordinates $[x_{\min}, y_{\min}, x_{\max}, y_{\max}]$.
    Segmentation MLLMs are designed for offline video understanding and do not provide a continuous streaming inference interface. 
    We therefore follow their offline evaluation setting.
    Besides, we finetune SA2VA-4B, UniPixel-3B and InstructSeg-3B on the train set of DroneEyes to compare with SkyAnchor-3B (denoted as the FT series model).
    \item Finally, to demonstrate our model's generality, we benchmark on the external SkyFind dataset against visual grounding models reported following the original benchmark paper~\cite{wang2026skyfind}. 
    These visual grounding model are training on the training set of SkyFind.
\end{itemize}
The general MLLMs in our study include closed-source (Gemini-3.0, GPT-5, GPT-4o, Claude-Sonnet-4.5, Kimi-K2.5) and open-source (Qwen-3.6, Gemma-4) models. 
For segmentation MLLMs, we evaluate the multimodal segmentation model SAM3, as well as MLLMs with inherent segmentation abilities (VISA, VideoGlaMM, InstructSeg, GLUS, VRS-HQ, SA2VA, and Unipixel).

\subsubsection{Training Strategy}
We adopt a progressive multi-stage training strategy. 
In Stages~1, we align the base vision language model backbone (Qwen2.5-VL-3B) and the segmentation projection head, and the SAM~2 decoder on a mixture of general-purpose referring segmentation datasets, establishing the model's core ability to perform language-guided segmentation on individual frames.
In Stage~2, we introduce the Token Router and train it jointly on our DroneEyes dataset. 
All modules are jointly optimized with learning rates.

\subsubsection{Real-world Deployment}
We deploy SkyAnchor on an NVIDIA Jetson AGX Orin 64GB for real-world edge inference.
Aerial videos are collected on the HKUST(GZ) campus in Guangzhou, China, using a DJI Matrice 4E UAV equipped with a 20-MP 4/3 CMOS wide-angle camera. 
The drone is flying at altitudes of $45$ m and $77$ m.

\begin{table}
\centering
\caption{Ablation study on components on SkyAnchor. }
\label{tab:ablation_of_components}
\resizebox{0.48\textwidth}{!}{
\begin{tabular}{@{}cc|ccc|ccc@{}}
\toprule
\multicolumn{2}{c|}{\textbf{Components}}                                                                                & \multicolumn{3}{c|}{\textbf{L1 - Object Description}} & \multicolumn{3}{c}{\textbf{L2 - Referring Expression}}              \\ \midrule
\begin{tabular}[c]{@{}c@{}}\textbf{Token}\\ \textbf{Router}\end{tabular} & \begin{tabular}[c]{@{}c@{}}\textbf{Memory}\\ \textbf{Bank}\end{tabular} & \textbf{ROUGE-L}        & \textbf{METEOR}        & \textbf{GPT}        & $\mathcal{J}$ & $\mathcal{F}$ & $\mathcal{J}\&\mathcal{F}$ \\ \midrule
                                                       &                                                       & 34.26           & 35.18         & 2.31       & 22.34         & 32.18         & 27.26                      \\
\cmark                              &                                                       & 37.52           & 38.64         & 2.56       & 24.17         & 35.60         & 29.89                      \\
                                                       & \cmark                             & 40.13           & 41.27         & 2.82       & 28.43         & 41.56         & 35.00                      \\
\cmark                              & \cmark                             & 45.40           & 46.31         & 3.22       & 30.96         & 45.23         & 38.10                      \\ \bottomrule
\end{tabular}
}
\end{table}

\begin{table}
\centering
\caption{Ablation study on Multi-layer Memory on SkyAnchor. }
\label{tab:ablation_of_memory}
\resizebox{0.48\textwidth}{!}{
\begin{tabular}{@{}cc|ccc|ccc@{}}
\toprule
\multicolumn{2}{c|}{\textbf{Memory Bank}}                                                                                      & \multicolumn{3}{c|}{\textbf{L1 - Object Description}} & \multicolumn{3}{c}{\textbf{L2 - Referring Expression}}              \\ \midrule
\begin{tabular}[c]{@{}c@{}}\textbf{Semantic}\\ \textbf{Memory}\end{tabular} & \begin{tabular}[c]{@{}c@{}}\textbf{Tracking}\\ \textbf{Memory}\end{tabular} & \textbf{ROUGE-L}        & \textbf{METEOR}        & \textbf{GPT}        & $\mathcal{J}$ & $\mathcal{F}$ & $\mathcal{J}\&\mathcal{F}$ \\ \midrule
                                                          &                                                           & 37.52           & 38.64         & 2.56       & 24.17         & 35.60         & 29.89                      \\
\cmark                                 &                                                           & 43.18           & 44.52         & 3.05       & 26.13         & 38.85         & 32.49                      \\
                                                          & \cmark                                 & 40.26           & 41.73         & 2.81       & 28.84         & 42.67         & 35.76                      \\
\cmark                                 & \cmark                                 & 45.40           & 46.31         & 3.22       & 30.96         & 45.23         & 38.10                      \\ \bottomrule
\end{tabular}
}
\end{table}

\begin{table}[t]
\centering
\caption{Comparison with other components using in existing MLLMs. }
\label{tab:comparison_with_components}
\resizebox{0.48\textwidth}{!}{
\begin{tabular}{@{}ccccccc@{}}
\toprule
\multicolumn{1}{c|}{\multirow{2}{*}{\textbf{Method}}} & \multicolumn{3}{c|}{\textbf{L1 - Object Description}}                   & \multicolumn{3}{c}{\textbf{L2 - Referring Expression}} \\ \cmidrule(l){2-7} 
\multicolumn{1}{c|}{}                                 & \textbf{ROUGE-L} & \textbf{METEOR} & \multicolumn{1}{c|}{\textbf{GPT}}  & \textbf{$\mathcal{J}$}        & \textbf{$\mathcal{F}$}       & \textbf{$\mathcal{J}\&\mathcal{F}$}   \\ \midrule
\multicolumn{7}{l}{\textit{\textbf{Token   Compression}}}                                                                                                                                \\ \midrule
\multicolumn{1}{c|}{Average   Pooling}                & 37.92            & 38.84           & \multicolumn{1}{c|}{2.58}          & 25.43             & 37.28            & 31.36           \\
\multicolumn{1}{c|}{Pixel-Unshuffle~\cite{bai2025qwen25vltechnicalreport}}                  & 40.16            & 41.05           & \multicolumn{1}{c|}{2.79}          & 26.88             & 39.42            & 33.15           \\ \midrule
\multicolumn{7}{l}{\textit{\textbf{Memory   Strategy}}}                                                                                                                                  \\ \midrule
\multicolumn{1}{c|}{SAM2 Native   Memory~\cite{ravi2024sam}}             & 42.74            & 43.68           & \multicolumn{1}{c|}{3.02}          & 27.06             & 39.74            & 33.40            \\
\multicolumn{1}{c|}{StreamingVLM   Memory~\cite{xu2025streamingvlm}}            & 43.55            & 44.47           & \multicolumn{1}{c|}{3.10}           & 28.12             & 41.18            & 34.65           \\ \midrule
\multicolumn{1}{c|}{\textbf{SkyAnchor   (Ours)}}      & \textbf{45.40}    & \textbf{46.31}  & \multicolumn{1}{c|}{\textbf{3.22}} & \textbf{30.96}    & \textbf{45.23}   & \textbf{38.10}   \\ \bottomrule
\end{tabular}
}
\end{table}

\begin{figure}
    \centering
    \includegraphics[width=1.0\linewidth]{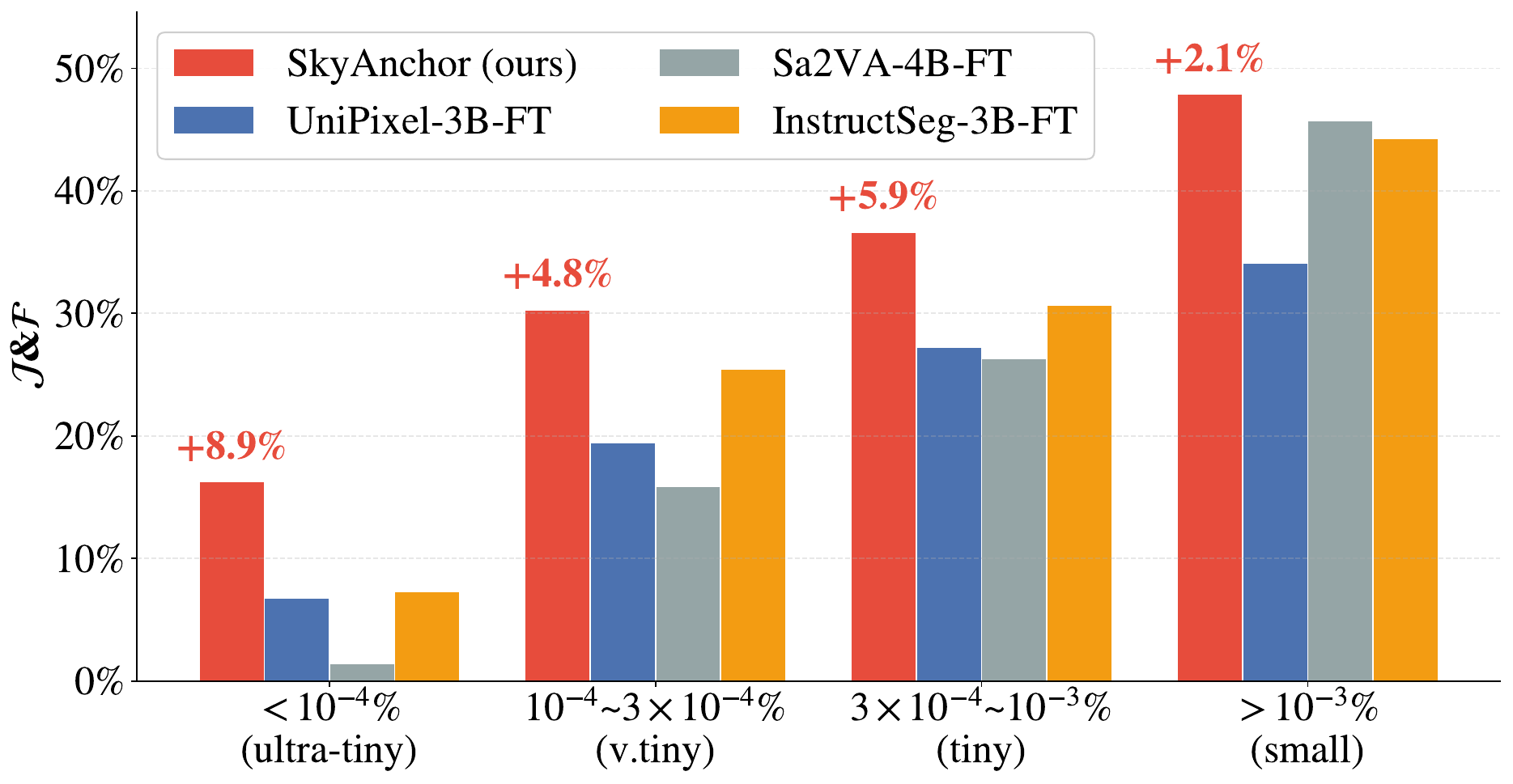}
    \caption{Per-size $\mathcal{J}\&\mathcal{F}$ comparison. 
    All three models are of comparable size (3B–4B) and fine-tuned on the same DroneEyes dataset. SkyAnchor outperforms three finetuned models across all size buckets, with a more margin on tiny objects. Numbers above bars denote the improvement of SkyAnchor over the best baseline.
    }
    \label{fig:size_bucket_chart}
\end{figure}

\subsection{Results on DroneEyes Video Dataset} 
Table~\ref{tab:l1} jointly reports the Level-1 object description and Level-2 referring segmentation results on DroneEyes.
\subsubsection{L1 Task}

Table~\ref{tab:l1} reports the Level-1 object description results. 
SkyAnchor achieves $45.40$ ROUGE-L, $46.31$ METEOR, and $3.22$ GPT Score with only 3B parameters, consistently outperforming all compared models.
General MLLMs struggle on this task. 
This shows that textual coordinate prompting is insufficient for precise spatial grounding in aerial imagery, where target objects are typically small and visually similar to surrounding context.
We attribute our advantage to the semantics-aware Token Router, which preserves fine-grained visual details of small objects during token compression. 
By dynamically assigning higher importance scores to target-relevant tokens, the Token Router retains discriminative features of small objects that would otherwise be smoothed out by uniform downsampling. 
This enables SkyAnchor to generate accurate attribute descriptions even for objects occupying a small portion of the aerial frame.

\subsubsection{L2 Task}

As shown in Table~\ref{tab:l1}, SkyAnchor (3B) achieves the best performance across all metrics, obtaining $\mathcal{J}$, $\mathcal{F}$, and $\mathcal{J}\&\mathcal{F}$ scores of $30.96 \%$, $45.23 \%$, and $38.10 \%$, respectively,  outperforming the second-best method SA2VA (26B, $\mathcal{J}\&\mathcal{F}$ = $33.05 \%$). 
General-purpose MLLMs including Gemini-3.0, GPT-5 and Claude-4.5 struggle to produce valid segmentation outputs on this task, indicating  limitations of generic multimodal models in fine-grained segmentation of small objects in aerial scenes. 
Among open-source segmentation-specific models, the Sa2VA series shows gradual improvement as model size increases (4B$\rightarrow$8B$\rightarrow$26B, $\mathcal{J}\&\mathcal{F}$ from $25.45 \%$ to $33.05 \%$), yet even the 26B variant falls far short of the 3B SkyAnchor, demonstrating that simply scaling up parameters cannot effectively address domain adaptation challenges in aerial imagery. 
Overall, SkyAnchor achieves state-of-the-art performance with the smallest parameter count, fully validating the effectiveness of its targeted design for small object referring segmentation in low-altitude aerial scenarios.

\begin{table*}[th]
\centering
\caption{Quantitative comparison with existing SOTA models in the SkyFind dataset to evaluate SkyAnchor generation. We report both macro-average and micro-average results.
Macro-average treats validation and test splits equally, while
micro-average weights each sample according to the split size.}
\label{tab:skyfind}
\begin{tabular}{@{}c|cccc|cccc@{}}
\toprule
\multirow{2}{*}{Method} & \multicolumn{4}{c|}{IoU@0.5}                                                           & \multicolumn{4}{c}{IoU@mean}                                                           \\ \cmidrule(l){2-9} 
                        & Val            & \multicolumn{1}{c|}{Test}           & Macro Avg.     & Micro Avg.     & Val            & \multicolumn{1}{c|}{Test}           & Macro Avg.     & Micro Avg.     \\ \midrule
FAOA~\cite{yang2019fast}                    & 23.10          & \multicolumn{1}{c|}{13.08}          & 18.09          & 15.41          & 10.21          & \multicolumn{1}{c|}{6.91}           & 8.56           & 7.68           \\
RSC~\cite{yang2020improving}                     & 29.61          & \multicolumn{1}{c|}{17.59}          & 23.60          & 20.38          & 16.38          & \multicolumn{1}{c|}{9.01}           & 12.70          & 10.72          \\
RefTR~\cite{li2021referring}                   & 31.22          & \multicolumn{1}{c|}{22.68}          & 26.95          & 24.66          & 15.80          & \multicolumn{1}{c|}{13.46}          & 14.63          & 14.00          \\
TransVG~\cite{deng2021transvg}                 & 35.49          & \multicolumn{1}{c|}{22.00}          & 28.75          & 25.13          & 20.56          & \multicolumn{1}{c|}{11.90}          & 16.23          & 13.91          \\
VLVTG~\cite{yang2022improving}                   & 30.29          & \multicolumn{1}{c|}{23.52}          & 26.91          & 25.09          & 14.40          & \multicolumn{1}{c|}{13.32}          & 13.86          & 13.57          \\
SeqTR~\cite{zhu2022seqtr}                   & 37.49          & \multicolumn{1}{c|}{25.74}          & 31.62          & 28.47          & 24.45          & \multicolumn{1}{c|}{12.57}          & 18.51          & 15.33          \\
QRNet~\cite{ye2022shifting}                   & 33.90          & \multicolumn{1}{c|}{26.21}          & 30.05          & 27.99          & 22.10          & \multicolumn{1}{c|}{11.22}          & 16.66          & 13.74          \\
SimREC~\cite{luo2023survivor}                   & 31.17          & \multicolumn{1}{c|}{21.50}          & 26.34          & 23.74          & 22.09          & \multicolumn{1}{c|}{12.15}          & 17.12          & 14.46          \\
PolyFormer~\cite{liu2023polyformer}              & 42.84          & \multicolumn{1}{c|}{31.01}          & 36.93          & 33.76          & 25.50          & \multicolumn{1}{c|}{16.44}          & 20.97          & 18.54          \\
AerialREC~\cite{wang2026skyfind}               & \textbf{45.21} & \multicolumn{1}{c|}{38.13}          & \textbf{41.67} & 39.77          & 29.78          & \multicolumn{1}{c|}{20.38}          & 25.08          & 22.56          \\ \midrule
SkyAnchor(ours)               & 32.66          & \multicolumn{1}{c|}{\textbf{48.02}} & 40.34          & \textbf{44.46} & \textbf{30.50} & \multicolumn{1}{c|}{\textbf{39.85}} & \textbf{35.18} & \textbf{37.68} \\ \bottomrule
\end{tabular}
\end{table*}

\subsection{Results on the Tiny Objects}

Figure~\ref{fig:size_bucket_chart} compares SkyAnchor with UniPixel-3B-FT and Sa2VA-4B-FT on the DroneEyes dataset, reporting $\mathcal{J\&F}$ across buckets defined by the target's relative area. The three models have comparable parameter counts (3B–4B) and were all fine-tuned (FT) on the same dataset. 
SkyAnchor consistently outperforms both baselines across all size buckets, and its advantage grows as the target size decreases: in the two smallest-target intervals, ultra-tiny (relative area $<10^{-4}\%$) and very-tiny ($10^{-4}\%\sim3\times10^{-4}\%$), SkyAnchor leads the strongest baseline by $9.5$ and $10.8$ percentage points, respectively. 
This suggests that SkyAnchor's design offers an advantage in small-target segmentation from a drone's perspective, rather than the improvement stemming merely from model scale or differences in fine-tuning data.

\subsection{Ablation Studies}

\subsubsection{Effectiveness of Components}

We examine the contribution of each proposed component by adding the Token Router and Memory Bank. 
As reported in Table~\ref{tab:ablation_of_components}, the base model without either component yields a $\mathcal{J}\&\mathcal{F}$ of $27.26 \%$, indicating that the plain VLM with SAM2 struggles with small object segmentation in aerial videos. 
Adding the Token Router alone improves $\mathcal{J}\&\mathcal{F}$ to $29.89 \%$, as it reduces background token interference and allows the model to attend more to foreground regions. 
Introducing the Memory Bank alone brings a  gain ($35.00 \%$), since temporal context across frames provides useful cues for locating and tracking small objects. 
When both components are combined, the model reaches $38.10 \%$ in  L2 $\mathcal{J}\&\mathcal{F}$ metrics, with consistent improvements on both L1 description metrics. 
This indicates that the two modules address complementary aspects of the problem. 
The Token Router improves spatial focus, while the Memory Bank supplies temporal information, and their combination yields the best overall performance.

\subsubsection{Comparison with Other Components}
Table~\ref{tab:comparison_with_components} presents a replacement-based comparison of the main components in SkyAnchor. 
In this table, we replace our Token Router with Average Pooling, a simple generic compression baseline, and Pixel-Unshuffle, the fixed-ratio token compression strategy used in Qwen2-VL and Qwen2.5-VL~\cite{bai2025qwen25vltechnicalreport}.
We also replace our Hierarchical Memory Bank with the native memory used in SAM2 and the streaming memory mechanism used in StreamingVLM~\cite{xu2025streamingvlm}. 
The results show that these alternative designs all lead to lower performance than SkyAnchor. This suggests that our Token Router is more suitable for preserving small-object information during token compression, while our Hierarchical Memory Bank is more effective at maintaining temporal context across frames.

\subsubsection{Ablation on Hierarchical Memory}
We further investigate the design of the Memory Bank by ablating its two layers: Semantic Memory and Tracking Memory. 
In this experiment, the Token Router is kept active in all configurations. 
As shown in Table~\ref{tab:ablation_of_memory}, without any memory layer the model achieves $29.89 \%$ $\mathcal{J}\&\mathcal{F}$, relying solely on per-frame processing. 
Adding Semantic Memory improves the L1 description quality noticeably (GPT score from $2.56$ to $3.05$), as it preserves object-level semantic features that help generate more accurate descriptions. 
Adding Tracking Memory instead leads to a  gain on L2 segmentation ($\mathcal{J}\&\mathcal{F}$ from $29.89 \%$ to $35.76 \%$), since it maintains spatial location history that directly benefits mask prediction. 
The full two-layer memory achieves the best results across all metrics, confirming that Semantic Memory and Tracking Memory serve different purposes and are both needed for the complete pipeline.

\subsubsection{Memory Capacity}

\begin{figure}
    \centering
    \includegraphics[width=0.8\linewidth]{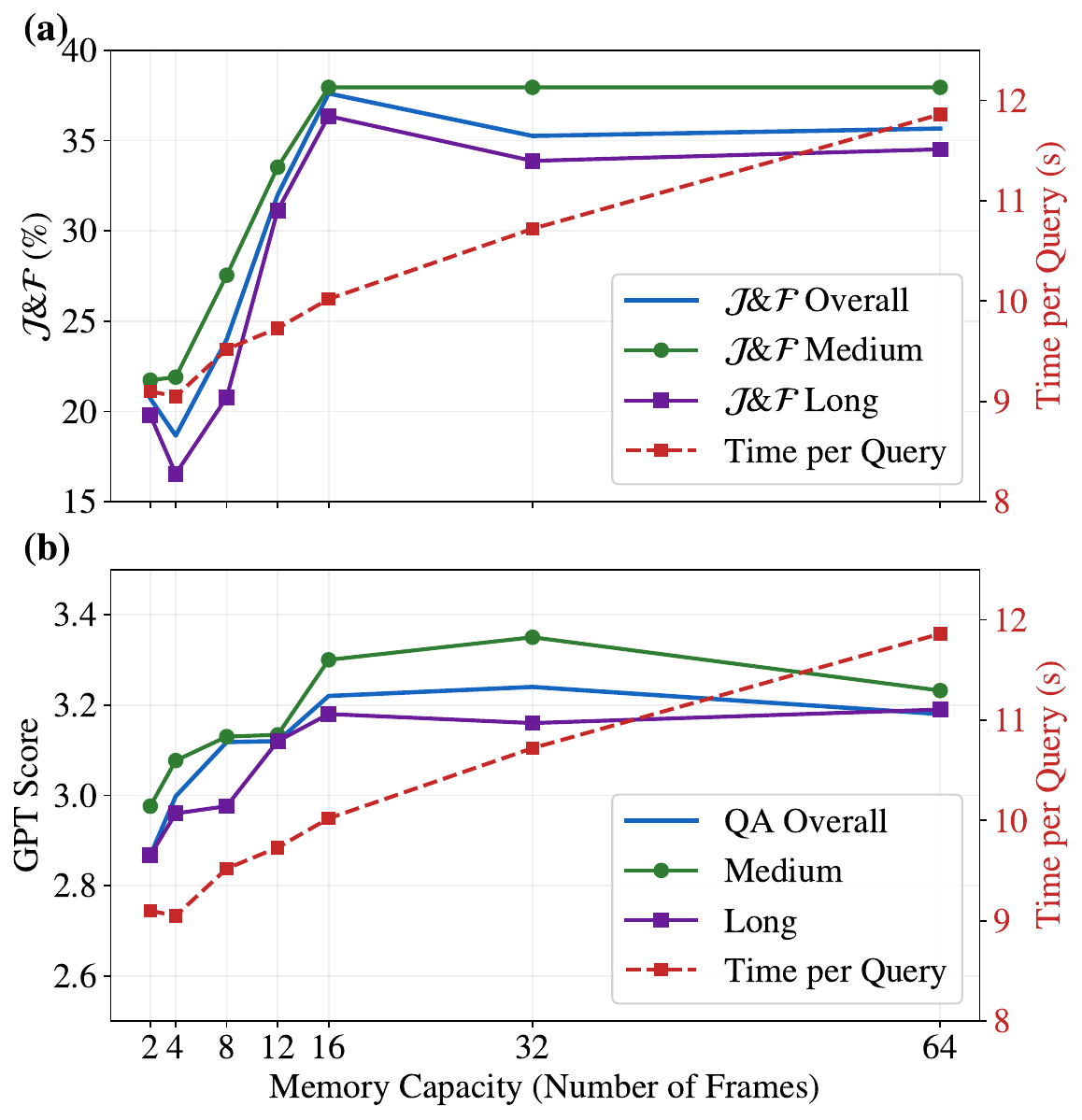}
    \caption{Ablation study on memory capacity. (a) Segmentation quality ($\mathcal{J}\&\mathcal{F}$) and inference time as a function of the number of memory frames.
    (b)  QA quality (GPT Score) and inference time as a function of the number of memory frames.
    }
    \label{fig:memory_capacity_ablation}
\end{figure}

As shown in Fig.\ref{fig:memory_capacity_ablation}(a), $\mathcal{J}\&\mathcal{F}$ improves sharply from $20.72\%$ ($2$ frames) and peaks around $16$ frames, after which adding more frames brings no further gain. 
Fig.\ref{fig:memory_capacity_ablation}(b) shows the same peak holds regardless of video length, but long videos ($>$ $500$ frames) degrade slightly beyond 16 frames due to attention dilution from stale context.
This indicates that more memory is not always better: $16$ frames already capture the effective temporal context.

\subsubsection{Token Routing}
\begin{figure}
    \centering
    \includegraphics[width=0.8\linewidth]{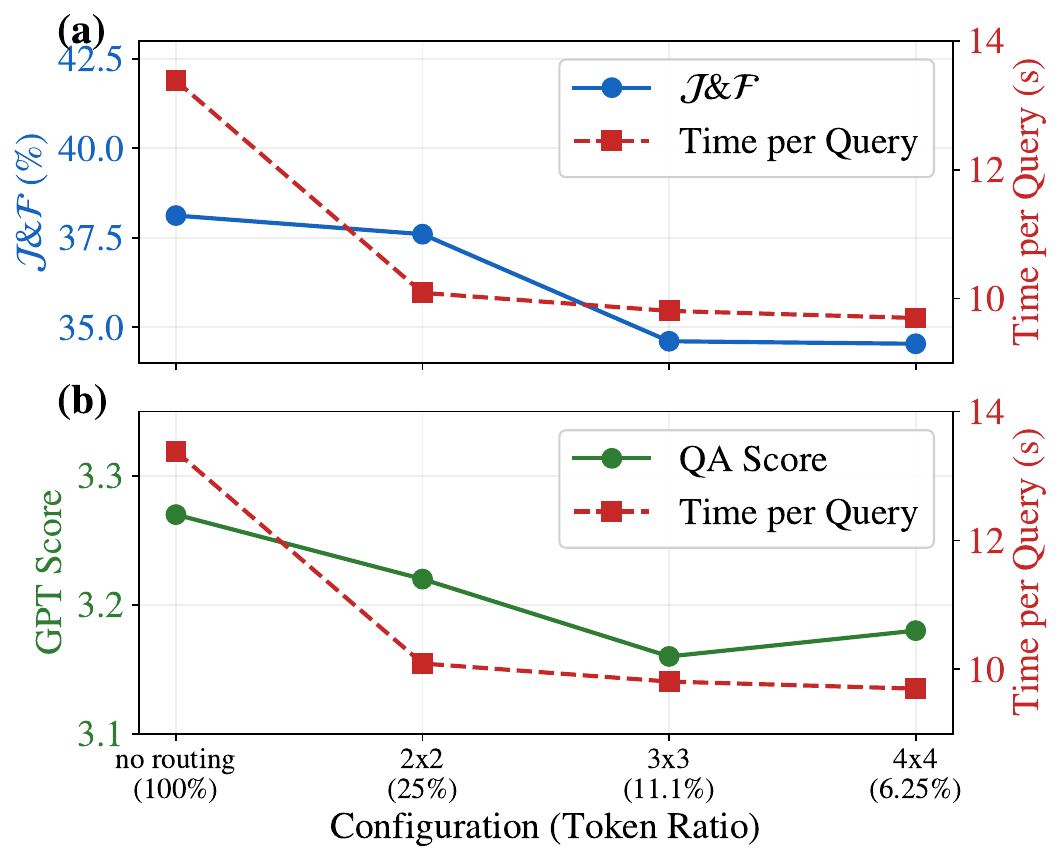}
    \caption{Ablation study on Token Router. (a) Segmentation quality ($\mathcal{J}\&\mathcal{F}$) and inference time under different setting.
    (b) QA quality (GPT Score) and inference time under different setting.
    }
    \label{fig:token_ablation}
\end{figure}

We conduct an ablation study on the token routing strategy to investigate the trade-off between model effectiveness and inference efficiency. 
As shown in Fig.~\ref{fig:token_ablation}, we compare four configurations: no routing (retaining $100 \%$ of visual tokens), and routing with $2 \times 2$, $3 \times 3$, and $4 \times 4$ spatial merging grids that reduce the token count to $25 \%$, $11.1 \%$, and $6.25 \%$, respectively. 
Without token routing, the model achieves a $\mathcal{J} \& \mathcal{F}$ of $38.12 \%$ but requires $13.38$ seconds per query averagely. 
Introducing the $2 \times 2$ token router reduces inference time by $24.6 \%$ (from $13.38$s to $10.09$s) with a performance drop of $0.02$ points in $\mathcal{J} \& \mathcal{F}$. 
The additional reduction in inference time is relatively small compared with the decrease in performance.
We therefore use $2\times2$ merging as the default setting.

\begin{figure}
    \centering
    \includegraphics[width=0.8\linewidth]{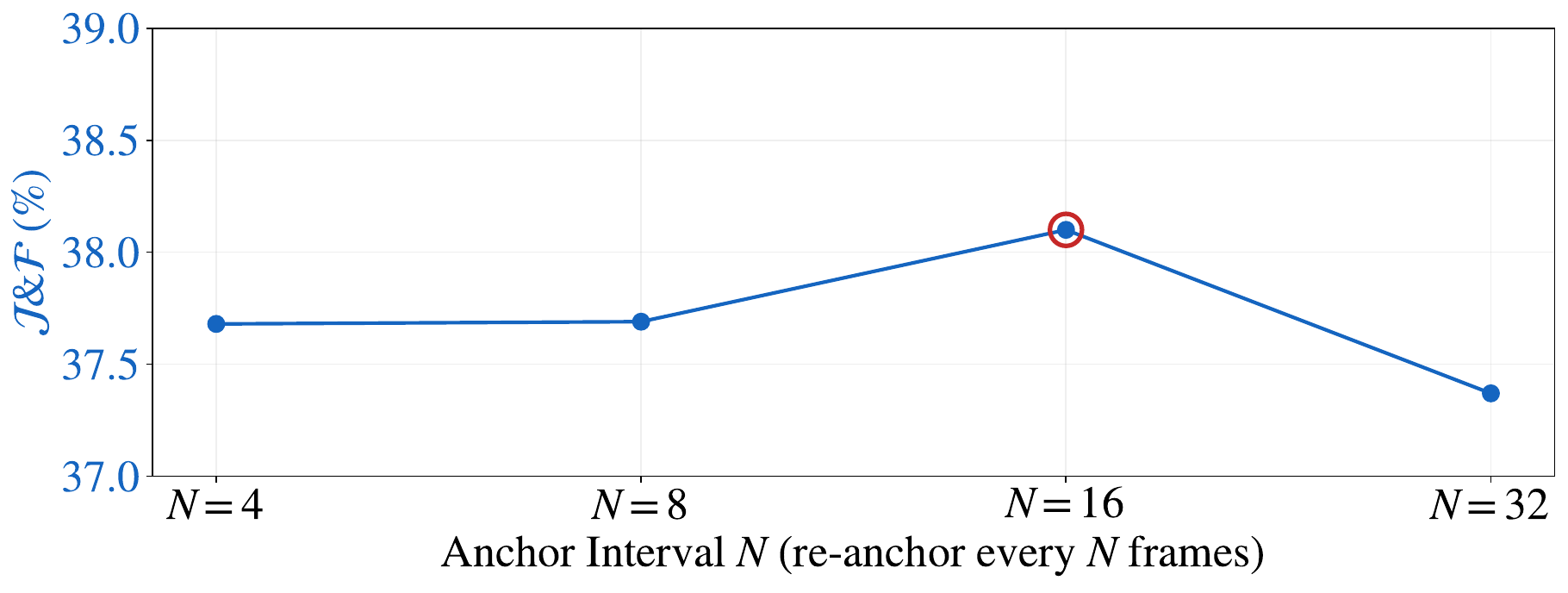}
    \caption{Ablation study of the anchor interval $N$ on SkyAnchor.
    }
    \label{fig:anchor_interval_ablation}
\end{figure}

\subsubsection{Ablation on Anchor Interval}
Figure~\ref{fig:anchor_interval_ablation} evaluates the effect of the
anchor interval $N$, which determines how often the VLM branch updates
the semantic anchor. The segmentation results vary within a limited
range for $N\in\{4,8,16,32\}$, with the highest score obtained at
$N=16$.
We use $N=16$ as the default setting because it gives the highest measured score while avoiding frequent VLM inference.

\subsection{Cross-Domain Generalization to Unseen Scene}
We benchmark SkyAnchor-3B against SOTA methods on SkyFind to evaluate  cross-domain generalization, as shown in Table~\ref{tab:skyfind}.
Unlike existing methods optimized on the SkyFind training split, SkyAnchor is evaluated without any additional training on this dataset.
On the unseen maritime scenes in the test set, SkyAnchor outperforms the SOTA AerialREC on both metrics, improving IoU@0.5 from $38.13$ to $48.02$ and IoU@mean from $20.38$ to $39.85$.
Under the micro-average evaluation, SkyAnchor achieves the best overall performance, obtaining $44.46$ IoU@0.5 and $37.68$ IoU@mean, compared with $39.77$ and $22.56$ for AerialREC, respectively.
Under the macro-average evaluation, which assigns equal importance to the validation and test splits, SkyAnchor achieves the best IoU@mean performance ($35.18$), while AerialREC obtains a slightly higher IoU@0.5 
score ($41.67$ vs. $40.34$).
This difference reflects the complementary strengths of the two models:
AerialREC performs strongly on the in-domain validation split, whereas SkyAnchor demonstrates superior robustness under the unseen domain shift.
These results indicate that models specifically optimized for the source domain may achieve stronger in-domain accuracy, while SkyAnchor provides more stable generalization when the distribution shifts.

\begin{figure*}
    \centering
    \includegraphics[width=0.8\linewidth]{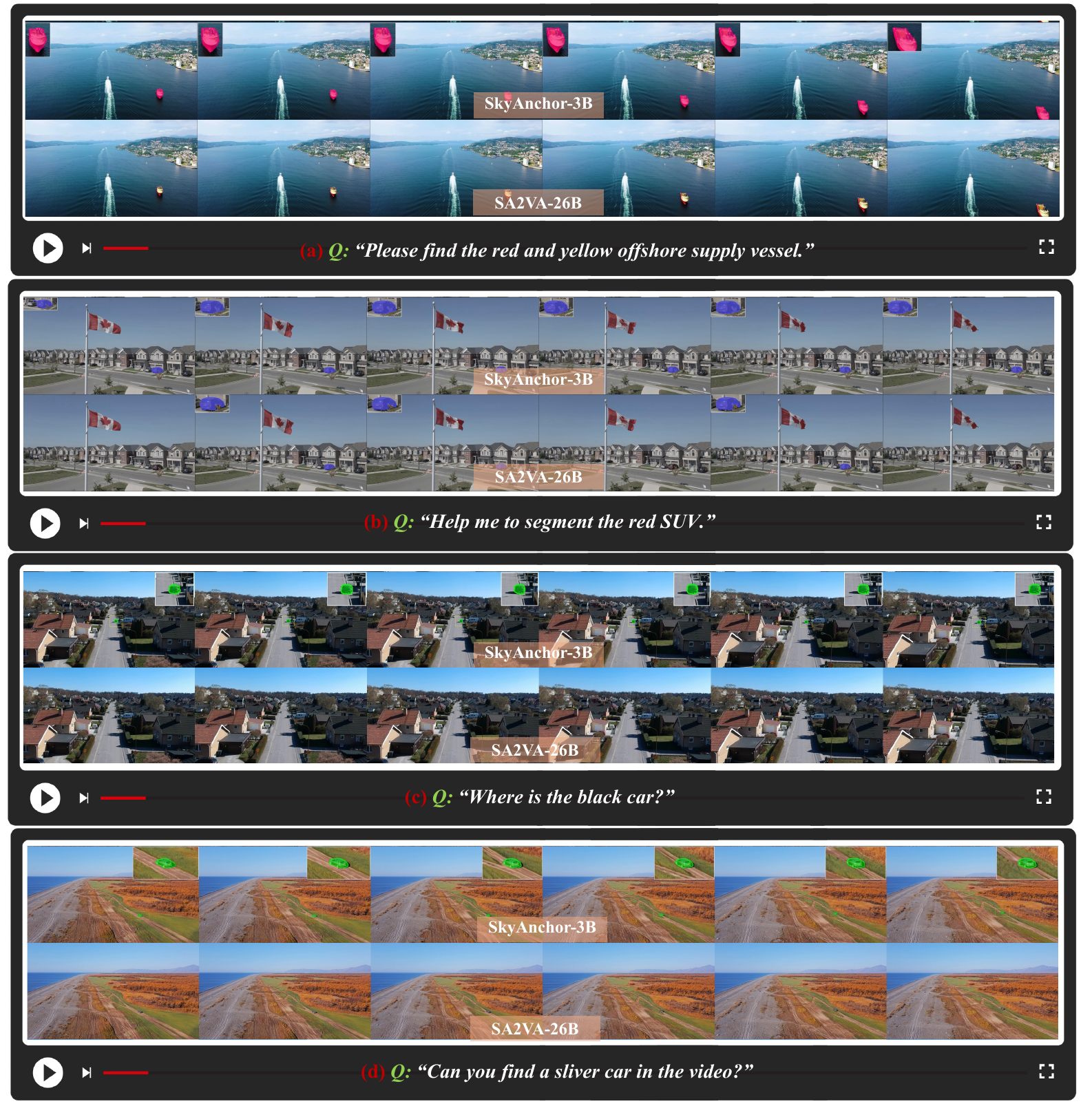}
    \caption{Qualitative results of SkyAnchor on four challenging video sequences. The input queries are shown in the green boxes. For each case, we display the result from SA2VA-26B (the best baseline model in Table~\ref{tab:l1}) and ours with zoomed-in insets for better visualization of details.
    In case (a), (c) and (d), SA2VA-26B cannot detect the target in the video. 
    In case (b), SA2VA-26B loses the target in some frames.
    }
    \label{fig:case_study}
\end{figure*}

\begin{figure*}
    \centering
    \includegraphics[width=0.8\linewidth]{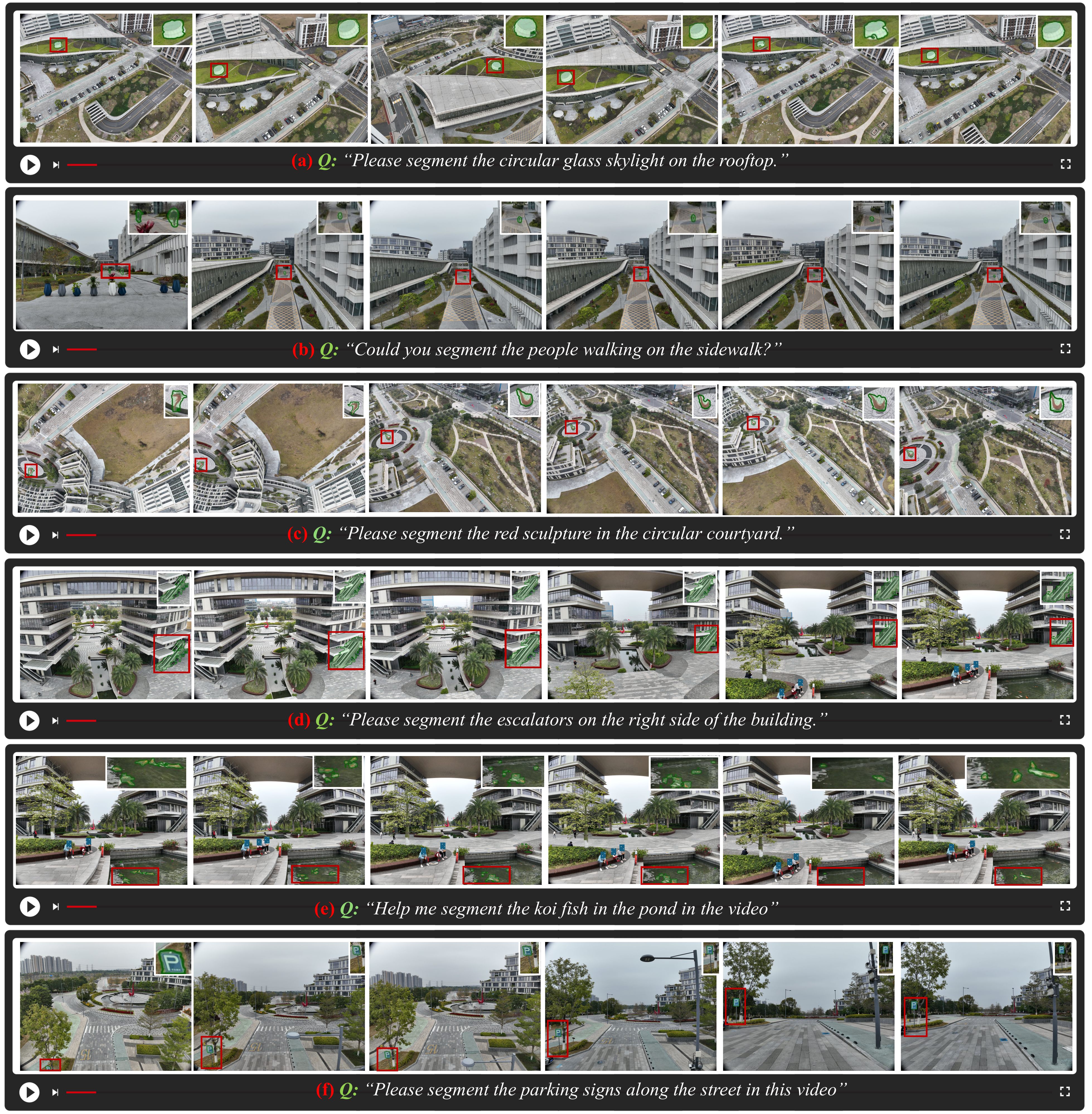}
    \caption{Qualitative results of SkyAnchor for the real world deployment. 
    The six queries are absent in the DroneEyes dataset.
    For privacy protection purposes, facial features in the video footage have been anonymized.}
    \label{fig:GKG_case_study}
\end{figure*}

\subsection{Device Deployment Analysis}

Table~\ref{tab:ablation_progressive_v2} reports the deployment efficiency of SkyAnchor on an NVIDIA Jetson AGX Orin 64GB device paired with a DJI Matrice 4E UAV. The vanilla PyTorch implementation achieves only $1.9$~FPS while consuming approximately $12.8$~GB of GPU memory. TensorRT compilation increases the throughput to $3.2$~FPS ($1.68\times$) and reduces memory usage to approximately $9.0$~GB through FP16 kernel fusion and reduced Python-side overhead. CUDA Graph capture of the SAM2 memory-attention path further improves the throughput to $3.7$~FPS ($1.95\times$), while asynchronous scheduling of the VLM and tracking branches delivers the largest gain, reaching $5.8$~FPS ($3.05\times$).

The optimized system produces a complete output every approximately $172$~ms, supporting cloud-free online monitoring for latency-tolerant applications such as aerial surveillance, infrastructure inspection, and operator assistance. Although it is not intended for high-speed closed-loop control or native-camera-rate processing, this throughput is consistent with aerial tracking and edge video analytics systems operating at low frame rates or invoking expensive perception modules only every few frames~\cite{yang2019multiple,ghosh2023react}. An SAM2-only configuration reaches $6.9$~FPS, or approximately $145$~ms per output, indicating that the remaining VLM-related overhead is only about $27$~ms, corresponding to $16\%$ of the complete pipeline latency.

\begin{table}
\centering
\caption{Ablation Study on Progressive Optimization for SkyAnchor Deployment.}
\label{tab:ablation_progressive_v2}
\begin{tabular}{l ccc}
\toprule
\textbf{Configuration} & \textbf{FPS} $\uparrow$ & \textbf{GPU Mem} $\downarrow$ & \textbf{Speedup} \\
\midrule
Vanilla PyTorch (Base) & 1.9 & {$\sim$12.8G} & 1.0$\times$ \\
\midrule
\multicolumn{4}{l}{\textit{TensorRT Backbone Optimization}} \\
\quad TensorRT & 3.2 & \multirow{4}{*}{$\sim$9.0G} & 1.68$\times$ \\
\quad + CUDA Graph Cache & 3.7 & & 1.95$\times$ \\
\quad + Async VLM & 5.8 & & 3.05$\times$ \\
\quad SAM2-Only (Upper Bound) & 6.9 & & 3.63$\times$ \\
\bottomrule
\end{tabular}
\end{table}


\subsection{Qualitative Results on DroneEyes}

Figure~\ref{fig:case_study} presents qualitative comparisons between SkyAnchor-3B and SA2VA-26B across four scenarios.
SA2VA achieves the second best results in the Table~\ref{tab:l1}.
In scenario (a), following the query \textit{``Please find the red and yellow offshore supply vessel,''} SkyAnchor-3B maintains tracking and segmentation as the vessel approaches from a distance, with relatively accurate boundaries under scale changes and textureless water surface. 
SA2VA-26B loses the target.
For scenario (b), with the query \textit{``Help me to segment the red SUV,''} SkyAnchor-3B segments the target vehicle in a residential street scene, distinguishing it from other parked cars without being affected by roadside trees and flags. 
SA2VA-26B detects the car in some frames but loses it in others. 
We attribute it to the memory usage in our SkyAnchor-3B.
In scenario (c), under the aerial-view query \textit{``Where is the black car?''}, SkyAnchor-3B localizes the small target spanning only a few pixels, without drifting to dark distractors such as shadows or roof textures. 
SA2VA-26B misses the vehicle in this bird's-eye view.
In scenario (d), answering \textit{``Can you find a silver car in the video?''}, SkyAnchor-3B separates the distant silver car from the visually similar gravel road in an open landscape, showing some ability to recognize color and texture differences. 
SA2VA-26B fails to identify the target in this low-contrast environment.
Across these examples, SkyAnchor shows three clear advantages. 
First, it maintains more stable tracking under large scale changes and long-range motion, as shown in the ship and red SUV cases. 
Second, it is more reliable for small targets, such as the black car that occupies only a few pixels in the aerial view. 
Third, it is less likely to drift to visually similar distractors, including shadows, roof textures, and low-contrast background regions.

\subsection{Qualitative Results on Realworld Deployment}

Figure~\ref{fig:GKG_case_study} shows the online segmentation results across different campus environments, guided by natural language queries\footnote{More video visualization demos can be seen in our website: \textit{\url{ https://sites.google.com/view/skyanchor/main-page}}}. 
The test cases cover six queries absent in the DroneEyes.
First, UAV maneuvers introduce continuous viewpoint changes: in cases (a) and (c), the circular skylight and sculpture are observed from varying oblique angles throughout the flight, yet the hierarchical memory bank preserves their structural semantics and keeps the masks stable without drifting to surrounding regions. 
Second, target scale varies dramatically across scenes. 
Case (d) involves large architectural structures such as building-side escalators, while cases (e) and (f) push to the opposite extreme where koi fish and street parking signs occupy only a handful of pixels. 
The semantics-aware Token Router selectively retains fine-grained cues for small targets while compressing uninformative background regions. 
Third, complex urban backgrounds introduce severe clutter and distractors. 
In case (b), pedestrians must be separated from standing plants using only the motion attribute in the query, and in case (c), the red sculpture must be grounded among other red-colored elements via a compositional spatial phrase. 
The MLLM leverages the full language query, including attribute, category, and spatial modifiers, to suppress distractors and maintain tracking continuity across frames. 
Across all six scenarios, SkyAnchor produces temporally coherent masks without any per-clip fine-tuning, demonstrating strong cross-domain robustness in real deployment.
These deployment results indicate that the model generalizes to different object categories and flight conditions.

\section{Conclusion}
\label{sec:conclusion}

This paper present DroneEyes, the \spl{first} large-scale aerial video referring segmentation dataset, and SkyAnchor, a memory-augmented MLLM framework for online referring segmentation in UAV video streams. 
DroneEyes addresses the absence of high-resolution aerial video benchmarks with dense pixel-level annotations by providing $2,140$ drone videos and $176,623$ grounding QA pairs organized into perception-description and referring-segmentation task levels. 
SkyAnchor tackles two core challenges of this domain: 
the Semantics-aware Token Router mitigates the extreme foreground-background imbalance by compressing irrelevant background tokens while preserving fine-grained target representations, and the Hierarchical Memory Bank decouples long-term semantic identity from short-term spatial tracking states. 
Extensive experiments show that SkyAnchor achieves state-of-the-art performance on both DroneEyes and the cross-domain SkyFind benchmark without fine-tuning.
Furthermore, we deploy SkyAnchor on the edge platform, achieving streaming inference without cloud offloading. 
Real-world experiments on campus surveillance videos validate the practical viability of the entire pipeline for online deployment.

Future work will pursue further model compression and hardware-aware optimization, enabling the pipeline to better match the native capture rate under fast target or platform motion.
Beyond efficiency, we will explore integrating active camera control and autonomous navigation with the segmentation pipeline, where the referring-segmentation output directly drives UAV attitude adjustment and path planning to adaptively keep the target within the field of view under an optimal scale and viewpoint.
We further plan to extend the framework toward multi-target referring as well as compositional queries involving relational and long-range temporal reasoning.

\bibliography{ref}
\bibliographystyle{IEEEtran}

\vspace{11pt}

\vfill

\end{document}